\documentclass[10pt,twocolumn,letterpaper]{article}

\usepackage{wacv}
\usepackage{times}
\usepackage{epsfig}
\usepackage{graphicx}
\usepackage{amsmath}
\usepackage{amssymb}
\usepackage{rotating}

\usepackage{multirow}
\usepackage{colortbl}
\usepackage{cite}
\definecolor{Gray}{gray}{0.9}
\newcolumntype{g}{>{\columncolor{Gray}} c}

\usepackage[pagebackref=true,breaklinks=true,letterpaper=true,colorlinks,bookmarks=false]{hyperref}

\wacvfinalcopy 

\def\wacvPaperID{704} 

\ifwacvfinal\fi
\begin{document}

\title{Quadtree Generating Networks: \\Efficient Hierarchical Scene Parsing with Sparse Convolutions}

\author{
{\centerline{
\begin{tabular}[t]{c}
Kashyap Chitta$^1$\thanks{Work completed while author is at CMU and NVIDIA.}\qquad Jos\'{e} M. \'{A}lvarez$^2$ \qquad Martial Hebert$^3$\\
$^1$ \textnormal{Autonomous Vision Group, MPI for Intelligent Systems and University of T\"{u}bingen}\\
$^2$ \textnormal{NVIDIA} \qquad $^3$ \textnormal{The Robotics Institute, Carnegie Mellon University}\\
{\tt\small kashyap.chitta@tue.mpg.de \qquad josea@nvidia.com \qquad hebert@cs.cmu.edu}
\end{tabular}
}}
}

\maketitle
\ifwacvfinal\thispagestyle{empty}\fi

\begin{abstract}
	Semantic segmentation with Convolutional Neural Networks is a memory-intensive task due to the high spatial resolution of feature maps and output predictions. In this paper, we present Quadtree Generating Networks (QGNs), a novel approach able to drastically reduce the memory footprint of modern semantic segmentation networks. The key idea is to use quadtrees to represent the predictions and target segmentation masks instead of dense pixel grids. Our quadtree representation enables hierarchical processing of an input image, with the most computationally demanding layers only being used at regions in the image containing boundaries between classes. In addition, given a trained model, our representation enables flexible inference schemes to trade-off accuracy and computational cost, allowing the network to adapt in constrained situations such as embedded devices. We demonstrate the benefits of our approach on the Cityscapes, SUN-RGBD and ADE20k datasets. On Cityscapes, we obtain an relative 3\% mIoU improvement compared to a dilated network with similar memory consumption; and only receive a 3\% relative mIoU drop compared to a large dilated network, while reducing memory consumption by over 4$\times$.
\end{abstract}

\begin{figure}
	\includegraphics[width=\columnwidth]{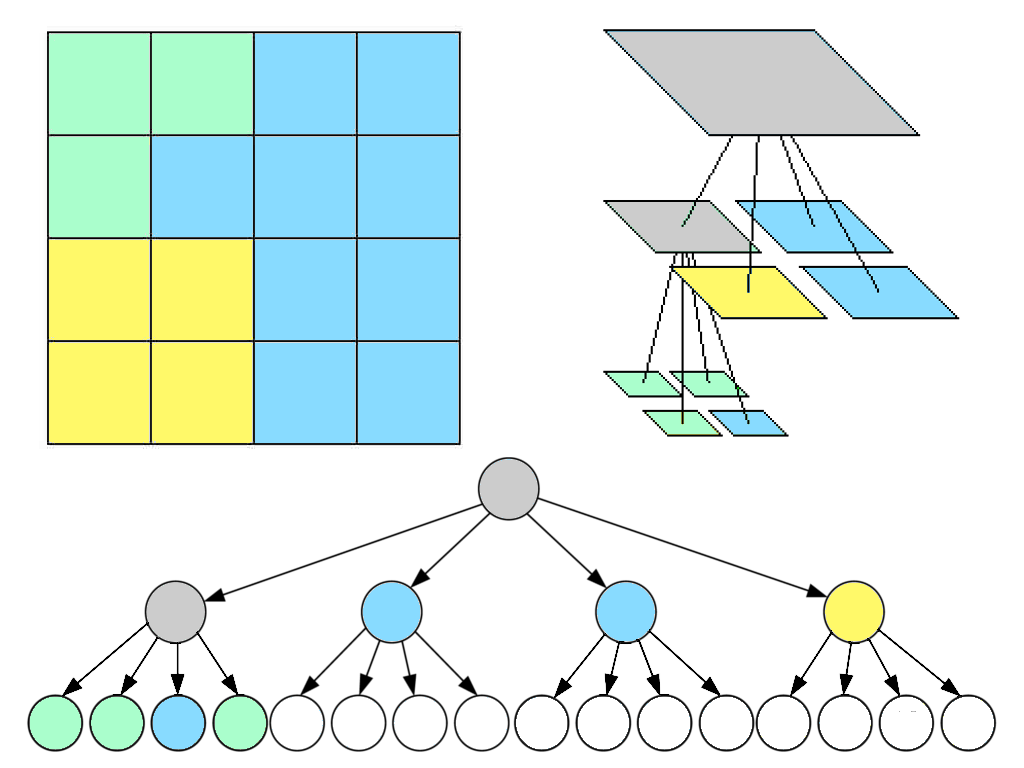}
	\caption{Efficient quadtree representation of a segmentation mask with three semantic classes (green, blue and yellow). Each node in the tree has four children, which represent quadrants of the corresponding image region in a clockwise order. The leaf nodes of the quadtree represent the pixel-level annotation. Some of the nodes in the quadtree (gray) are composite, and must be further split into individual classes; whereas others (white), whose parents in the quadtree structure are not composite, can be removed from the representation, achieving lossless compression (best viewed in color).}
	\label{f:quadtree}
\end{figure}

\section{Introduction}
Semantic segmentation is the problem of assigning a class to every pixel in an image. It is a challenging task as it requires a combination of coarse contextual reasoning and fine pixel-level accuracy \cite{he2004multiscale,farabet2013learning}. Fully Convolutional Networks (FCNs) \cite{long2015fully} and their variants \cite{ronneberger2015unet,noh2015learning,zhao2017pspnet,chen2017rethinking} have become a standard tool for this task, leveraging the representational power of deep Convolutional Neural Networks (CNNs) trained end-to-end with backpropagation. In contrast to traditional methods using low-level features \cite{achanta2012slic}, FCN-based approaches extract information at various scales using pooling and subsampling layers to perform contextual reasoning and obtain significant accuracy improvements. However, a common limitation of all these FCN-based methods is that they build upon convolutional layers, which require dense pixel grids at their input and output. Compute and memory requirements for dense pixel grids scale quadratically with the resolution of the input image, thus increasing the inference time required to process high-resolution images~\cite{romera2018erfnet}. In addition, at train time, this large memory requirement limits the number of samples and spatial resolution used in the training batch. Due to this limitation, most recent architectures can fit just two cropped images per GPU while training \cite{zhao2017pspnet,chen2017rethinking}. Existing tricks to circumvent this such as CPU/GPU activation swapping introduce significant overhead in training times.

In this paper, we propose a novel encoder-decoder based segmentation approach to reduce memory consumption in the most computationally demanding layers of a segmentation network. Our key idea is to decompose the target segmentation mask into a linear quadtree representation, as shown in Fig. \ref{f:quadtree}. The use of quadtree representations enables us to replace several dense convolutional layers of the network's decoder with sparse convolutions \cite{ren2018sbnet,graham2018semantic} that are significantly more efficient than their dense counterparts, and scale linearly in terms of compute and memory requirements to higher resolutions. In our experiments, the quadtree representation of the high-resolution labels used consumes $20\times$ to $30\times$ less memory than the corresponding dense pixel-wise representation. Further, our proposed efficient and high-performance decoder enables the use of less computationally demanding backbone networks. Overall, our networks are $2\times$ to $4\times$ more efficient than strong baselines, with similar or better performance.

In short, our contributions are: (i) a novel approach to use quadtrees to represent semantic segmentation masks for deep neural networks; (ii) a novel architecture, Quadtree Generating Networks (QGNs), able to generate quadtrees from input images efficiently using sparse convolutional layers; and (iii) a loss function to train QGNs using high-resolution images. We demonstrate the benefits of our approach on three public semantic segmentation datasets. Our results consistently outperform existing networks with similar memory requirements and perform similarly to current state-of-the-art models while significantly reducing the memory and computational requirements. Moreover, at inference time, our approach is flexible, and allows a trade-off to be made between accuracy and computational requirements without any modification of a trained network.

\section{Related Work}

\noindent
\textbf{Scene Parsing.} Semantic segmentation of complex scenes has seen considerable progress in recent years through FCN-based methods. Recent approaches fall under three broad categories: methods to improve contextual reasoning, methods to improve feature resolution, and methods that allow fast inference while maintaining reasonable performance. The first category of works propose modules, typically incorporated at the final layers of the encoder, to augment the feature representation and improve contextual reasoning. Though the receptive field of deep CNNs is theoretically large, studies have shown that their effective receptive field is small \cite{luo2016understanding}. Atorus Spatial Pyramid Pooling (ASPP) in DeepLabv3 \cite{chen2017rethinking} and the Pyramid Pooling Module (PPM) in PSPNet \cite{zhao2017pspnet} are two modules which apply kernels of various large scales to the feature map-- ASPP employs dilated convolutions whereas PPM uses average pooling for capturing global context. Scale Adaptive Convolutions (SAC) \cite{zhang2017scale} achieve a similar effect by modifying the convolutional operator to build larger receptive fields. More recently, PSANet \cite{zhao2018psanet}, DANet \cite{fu2018dual} and CCNet \cite{huang2018ccnet} encode context using self-attention mechanisms \cite{vaswani2017attention,wang2017nonlocal}.

The second category of work focuses on the performance at boundaries between classes and other fine details in images, which is directly tied to feature resolution. Typically, pre-trained classification architectures give feature maps that are 32 times smaller than the input resolution. Early approaches on improving this aspect of segmentation include the use of skip connections \cite{long2015fully}, deconvolutional architectures \cite{noh2015learning,badrinarayanan2017segnet,lin2017refinenet}, or leveraging the features learned at all layers in the network to make predictions \cite{hariharan2015hypercolumns,ronneberger2015unet,pinheiro2016learning}. Since then, dilated convolutions have become the standard technique to handle the resolution issue, by maintaining a fairly high feature resolution throughout the network, while also increasing the receptive field as a subsampling or pooling layer would \cite{yu2016dilated}. State-of-the-art segmentation networks based on the ResNet encoder architecture \cite{he2016deep} typically dilate the last two residual blocks, giving output feature maps that are only 8 times smaller than the input resolution. However, given the design of ResNets, it becomes a huge computational burden to extract output feature maps at this scale. For example, with the ResNet-101 architecture, \textit{78 of the 101 layers} would now have 4 or 8 times greater memory consumption.

The third category includes work on designing decoders and other efficient alternatives to dilation \cite{chen2018encoder,xiao2018upernet}. However, the quadratic scaling of standard operations used in these techniques, such as deconvolution and bilinear upsampling, remains a crucial issue, which has seen little to no study in recent semantic segmentation literature \cite{wojna2018devil}. In the proposed approach, we focus on this issue. Rather than architectural modifications, we aim to achieve better scaling through a novel representation of the network outputs.

\noindent
\textbf{Quadtree and octree representations.} A quadtree is a hierarchical data structure with a long history in image processing literature \cite{samet1984quadtree,sonka2007image}. Quadtrees were prominently used in some of the earliest work on semantic segmentation with graphical models, due to their hierarchical structure and efficiency \cite{bouman1994multiscale,feng1998training}. In contrast, quadtrees in the deep learning era have only recently been applied, in a very limited setting, for tasks involving sparse, binary input images \cite{jayaraman2018quadtree}. Our proposed networks, on the other hand, do not make any sparsity assumptions on their inputs.

The family of octree representations, which are the 3D analog of quadtrees, has had significant impact on 3D deep learning. In this domain, sparse convolutions on octrees alleviate the burden of cubic scaling \cite{riegler2017octnet}. Our work is closely related to Octree Generating Networks, which perform high resolution 3D reconstruction from single images using a deconvolutional architecture \cite{tatarchenko2017octree}. The key differences are: (i) a novel, flexible loss formulation, that allows the user to trade-off between accuracy and computational cost; (ii) deeper decoder architectures and more quadtree levels than those typically used with octrees; (iii) inclusion of elements such as encoder-to-decoder skip connections and adaptive level-wise loss weighting; and (iv) quadtree representations over semantic labels of complex scenes with a large number of semantic classes (up to 150); as opposed to work in the 3D setting, which has largely been restricted to binary representations (such as voxel occupancy grids).

\noindent
\textbf{Anytime inference.} Anytime predictors are a class of algorithms that produce a crude initial prediction for each test sample, and continue to refine it as allowed by a compute budget \cite{grass1996anytime,hu2016efficient}. There are several real-world situations where inference-time computational costs of scene parsing are critical, which prompted research into anytime predictors for semantic segmentation in the era of fixed feature extractors \cite{grubb2013speedmachines,liu2016learning}. However, anytime prediction with learned CNN based feature extractors has been evaluated primarily in the image classification setting \cite{hunag2017multiscale,hu2017learning,lee2018anytime}. In contrast, our work with QGNs involves anytime inference for semantic segmentation, through the inherent flexibility of the quadtree representation.

\section{Quadtree Generating Networks}
\label{s:QGN}
\begin{figure*}[t]
	\includegraphics[width=\textwidth]{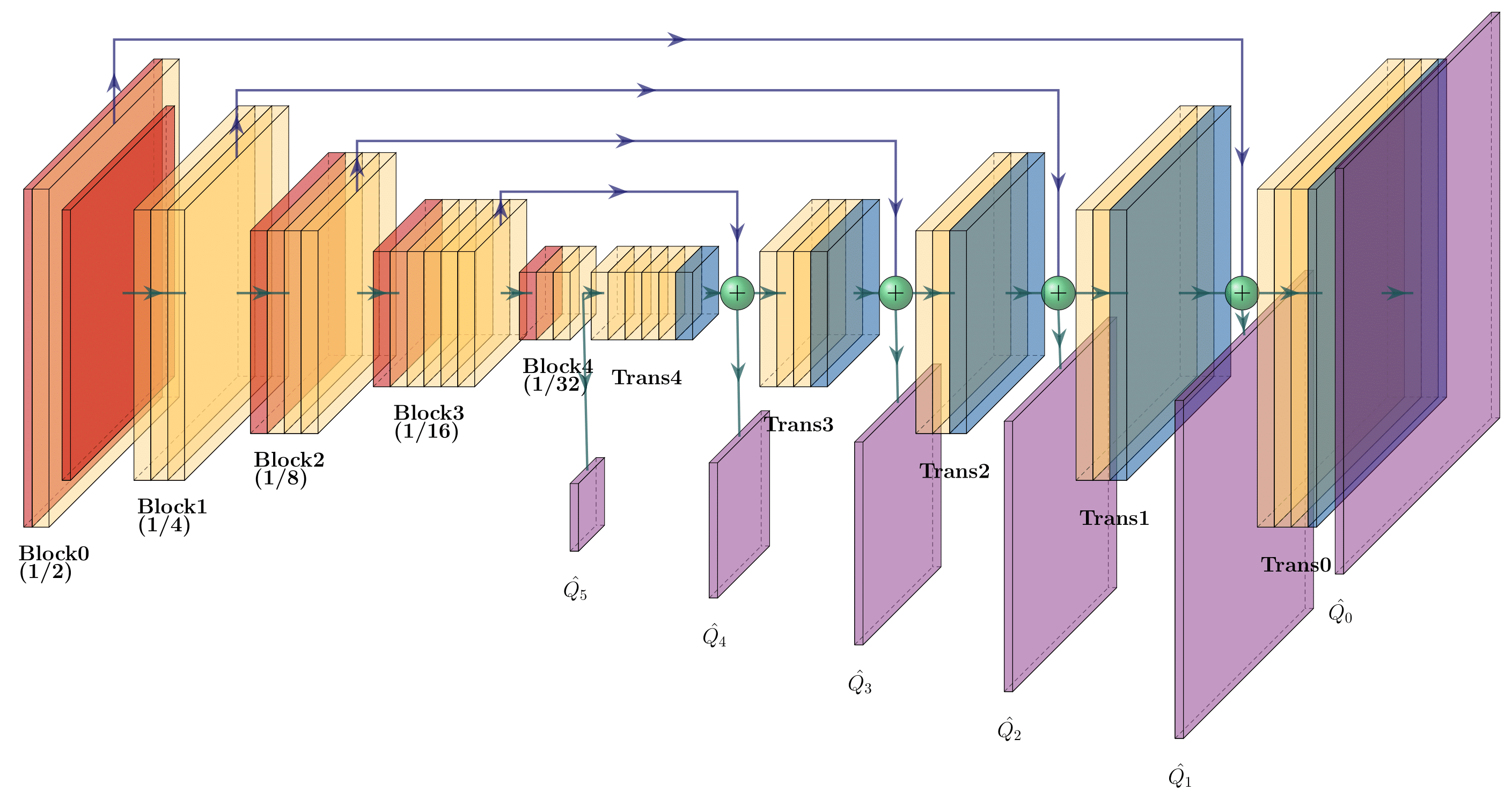}
	\caption{QGN network architecture, consisting of an encoder, decoder, skip connections and prediction layers. Subsampling layers in the encoder are shown in red, and upsampling layers in the decoder in blue. Prediction layers with the softmax activation are shown below the main architecture in purple. The first such prediction layer is placed immediately after the encoder, for which the output $\hat{Q}_5$ is of resolution $\frac{1}{32}$ compared to the input, and forms the root node of the quadtree. Prediction layers after each skip connection the decoder produce additional quadtree levels $\hat{Q}_4, ..., \hat{Q}_0$ (best viewed in color).}
	\label{f:arch_fig}
	\vspace{-0.5cm}
\end{figure*}

In this section, we describe our deep network architecture used to generate an efficient quadtree prediction for a given input image, the Quadtree Generating Network (QGN). Its architecture is shown in Fig.~\ref{f:arch_fig}. We begin by introducing the quadtree and T-pyramid representations, used in calculating the loss function for training QGNs. We then describe the three main components of our architecture: the sparse convolutional decoder, prediction layers and skip connections. We limit our focus in this study to the setting where the encoder subsamples inputs by a factor of 32. Correspondingly, we use a sparse convolutional decoder with 5 blocks-- each block consists of several sparse convolutions followed by an upsampling operation.

\subsection{Quadtrees and T-Pyramids}
\label{s:quadtree}

A quadtree is a tree data structure in which each internal node has exactly four children \cite{samet1984quadtree}, see Fig.~\ref{f:quadtree}. Quadtrees are the two-dimensional analog of octrees and are most often used to partition a two-dimensional space by recursively subdividing it into four quadrants or regions. A function defined on a pixel grid, such as a segmentation map, can be converted to a quadtree by this kind of recursive subdivision of cells-- if every pixel within a sub-cell has the same function value, it need not be further divided, and becomes a leaf of the quadtree assigned that specific function value. 

The set of cells at a certain resolution in a quadtree is referred to as a quadtree level, denoted by $l$. If a fixed number of levels is desired, the recursive subdivision process can be started not from the whole grid, but from some initial coarse resolution. Then, the maximal quadtree cell size, or size of the root node, is given by this initial resolution. For example, a maximal cell size of $32 \times 32$ can be partitioned into a quadtree with 6 levels.

An efficient way of storing a quadtree is using hash tables: each node in the quadtree is indexed by its key, which is a tuple of the level and spatial location $(l,x,y)$, and stores a value $v$, which may be discrete or continuous. In the case of semantic segmentation, the value $v$ comes from the set $\{1,...,k\}$ where $k$ is the number of classes. The quadtree $Q$ is the set of all key-value pairs,
\begin{equation}
    Q = \{(l,x,y,v)\}.
\end{equation}

To train networks using quadtree representations, we would need to be able to compute the distance between two different quadtrees, the prediction $\hat{Q}$ and ground truth $Q$. To this end, we would need to be able to query, for any cell $(l,x,y)$ in $\hat{Q}$, the corresponding value in $Q$. This is not trivial, as the two quadtrees may have a different structure. The ground truth value corresponding to a prediction with the wrong structure would be undefined. To resolve this discrepancy, we store the ground truth $Q$ in the a more generalized form, called a tree-pyramid (T-pyramid), denoted as $T$. A T-pyramid is a 'complete' quadtree; every node of the T-pyramid has four children except leaf nodes; and all leaves are on the same level, which corresponds to individual pixels in the image \cite{sonka2007image}. To efficiently construct this kind of representation for a segmentation mask with classes $(1, ..., k)$, we propose the use of a merging operator $\mathcal{M}$ that is applied to patches of size $2\times 2$,
\begin{equation}
    \mathcal{M} \left( \begin{bmatrix}
    a & b \\
    c & d
    \end{bmatrix} \right) = 
    \begin{cases}
    a,& \text{if } a=b=c=d\\
    0,              & \text{otherwise,}
\end{cases}
\label{e:merge_op}
\end{equation}

\noindent
where $0$ is the value assigned to a newly introduced class that represents a mixture of multiple classes. We refer to this as the 'composite' class (gray cells in Fig. \ref{f:quadtree}). This class helps to encode information about the quadtree structure in the T-pyramid representation, as any cell belonging to this class has children at a lower level in the original quadtree. Cells with any value other than the composite class have no children, or are not present in the original quadtree.

Let $T_0$ represent the level 0, or leaf nodes, of the T-pyramid. $T_1=\mathcal{M}(T_0)$ is obtained by applying $\mathcal{M}$ to all the $2\times 2$ patches in $T_0$. Similarly, we can continue building the pyramid to as many levels as desired, by recursively applying $\mathcal{M}$,
\begin{equation}
    T_{l+1}=\mathcal{M}(T_l).
    \label{e:t_pyr}
\end{equation}

\noindent
\textbf{Loss function.} We now formulate the loss function between a quadtree prediction $\hat{Q}$, and the corresponding ground truth T-pyramid label $T$. At each level $l$ in $\hat{Q}$, the level loss $\mathcal{L}_l$ is computed as:

\begin{equation}
    \mathcal{L}_l = \frac{1}{N} \sum_{i=1}^{N} \mathcal{H}(v^i,\mathcal{T}(l,x^i,y^i)),
    \label{e:loss_l}
\end{equation}

\noindent
where $\mathcal{H}$ is the cross-entropy loss function over the $k+1$ classes that may be taken by $v^i$ (composite + $k$ actual classes). $\mathcal{T}$ is a query function on the T-pyramid that returns the value of the corresponding hash table key,
\begin{equation}
    \mathcal{T}(l,x,y) = v.
\end{equation}

The overall loss is a weighted sum of the loss at each level, given by:
\begin{equation}
    \mathcal{L} = \sum_{l \in Q} \beta_l \mathcal{L}_l.
    \label{e:loss_full}
\end{equation}

\noindent
\textbf{Loss weighting.} We experiment with two approaches for weighting the loss at different levels of the quadtree (Eq.~\ref{e:loss_full}): (i) \textbf{Fixed}, where we scale the weight assigned to each layer by a multiplicative hyper-parameter $\gamma$ with respect to the preceding layer's weight (with $\beta_0=1$),
\begin{equation}
    \beta_{l+1} = \gamma \beta_{l};
    \label{e:fix_loss}
\end{equation}

\noindent
and (ii) \textbf{Adaptive}, where we use a running average value of the loss at that scale as a weight, as in \cite{hu2017learning}:
\begin{equation}
    \beta_{l}^{i+1} = \delta \beta_{l}^{i} + (1-\delta)\mathcal{L}_l^i,
    \label{e:adapt_loss}
\end{equation}

\noindent
where $i$ is the training iteration index, each $\beta_l$ is initialized to 1, and $\delta=0.99$ in our experiments.

\noindent
\textbf{Sparse activations.} To compute $\mathcal{L}_l$ as defined in Eq. \ref{e:loss_l}, the desired output from our network architecture is a quadtree prediction $\hat{Q}$. Our decoder generates the entries of $\hat{Q}$ in a level-by-level manner.

Let the network activation $A_l$ have a resolution corresponding to the quadtree level $\hat{Q}_l$. For example, after passing through a typical state-of-the-art deep encoder, which subsamples the input by a factor of 32, the activations $A_5$ would correspond to the quadtree level $\hat{Q}_5$.

$A_l$ is typically stored as a tensor of dimensions $W_l \times H_l \times C_l$. However, in our decoder, we store the activations as hash tables,
\begin{equation}
    A_l=\{(x,y,v)\},
    \label{e:sparse_act}
\end{equation}

\noindent
where $x \in \{1,...,W_l\}$, $y \in \{1,...,H_l\}$, and $v \in \mathbb{R}^{C_l}$. Note that, the T-pyramid labels (Eq. \ref{e:t_pyr}) are quadtrees, whose values $v$ are class indices. On the other hand, sparse activations (Eq. \ref{e:sparse_act}) are hash tables, whose values $v$ are float numbers identical to those used in the tensor representation.

\subsection{Architecture Components}

\noindent
\textbf{Sparse convolutions.} Sparse convolutions are the fundamental building blocks of our decoder. For any sparse activation $A_l$ (Eq. \ref{e:sparse_act}), we define $a$ to be a set of \textit{active sites}, which can be \textit{any subset} of all spatial locations in $A_l$. A sparse convolutional layer with kernel $W$, bias $b$ and active sites $a$ operates on only the set of values at these active sites, generating an new activation map $A_l'=(x,y,v')$ with the same active sites as the input, through a transformation to its values:

\begin{equation}
    v' = 
    \begin{cases}
    Wv + b ,& \text{if } (x,y) \in a\\
    0,              & \text{otherwise.}
\end{cases}
\end{equation}

Using this procedure, multiple sparse convolutional layers can be used to propagate activations through the network in hash table form, without changing the active sites \cite{graham2015sparse}.

\noindent
\textbf{Upsampling.} After a set sparse convolutions, each block in our decoder is differentiated from the next block through an upsampling operation. Upsampling changes the resolution, and hence the quadtree level of the activation, from $l$ to $l-1$. We perform this operation through $2\times 2$ nearest neighbor interpolation, an operation that doubles the resolution of the activation (i.e., $W_{l-1} = 2W_l$ and $H_{l-1} = 2H_l$). $A_{l-1}$ is obtained as follows:

\vspace{-0.5cm}
\begin{multline}
    A_{l-1} = \{ (2x,2y,v),
    (2x+1,2y,v), 
    (2x,2y+1,v), \\
    (2x+1,2y+1,v)\} \ \forall \ (x,y,v) \in A_l.
    \label{e:upsample}
\end{multline}

\noindent
\textbf{Skip connections.} To allow the deeper layers in the decoder to access high-frequency information from early in the encoder, otherwise lost due to subsampling, we incorporate additional $1 \times 1$ convolutional skip connections from the encoder to its corresponding block in the decoder, as shown at the top of Fig. \ref{f:arch_fig}. Skip connections are essential for a decoder with our choice of upsampling operation, since the upsample described in Eq. \ref{e:upsample} produces outputs $A_{l-1}$ that have identical values in local $2\times2$ blocks at the finer resolution. By adding non-identical features from the encoder to the identical features coming from the previous block in the decoder, we obtain a unique feature value at each spatial resolution in $A_{l-1}$.

\noindent
\textbf{Prediction layers.} A $1 \times 1$ convolutional prediction layer is used to generate $\hat{Q}_{l}$ from the activation $A_{l}$ (downward arrows in Fig. \ref{f:arch_fig}). Prediction layers use a softmax activation over $k+1$ channels, corresponding to the $k$-way classification problem in the dataset, and the additional composite class. At train time, the level-wise loss from Eq. \ref{e:loss_l} is computed at each of these prediction layers. Inference at test-time is conducted by accumulating the predictions made at the leaf nodes of $\hat{Q}$. In case the composite class has the highest confidence at any leaf node, the class with second-highest confidence is assigned as the prediction at that pixel instead.

\subsection{Propagation Schemes}
\label{ss:prop}

An important component of the optimization is the choice of active sites $a$ that are propagated at each level of the decoder, which we call the propagation scheme. The choice of propagation scheme is extremely flexible-- for example, certain high-priority classes could be propagated, while efficiency is maintained by not propagating entries assigned to other classes. As another use case, an entire block of the decoder could be left unpropagated by choosing no active sites, when fine detailed segmentation is not necessary. In this work, we focus our attention on three different propagation schemes: (i) All; (ii) GT Composite; and (iii) Predicted Composite.

\noindent \textbf{All.} All pixels are propagated through the complete decoder till the leaf nodes. Serves as a baseline with the maximum possible memory consumption, and can be used during both training and inference, given sufficient resources.

\noindent \textbf{Ground Truth Composite (GTC).} The quadtree representation of the ground truth segmentation mask is used to decide which pixels to propagate. \textit{Only pixels of the composite class in the ground truth} are selected as active sites at any given layer in the network. Due to the sparsity in ground truth, this leads to significantly reduced memory consumption. In practice, this mode of operation is useful when there is access to the ground truth (during training) but cannot be used for inference. However, in our experiments, we include results for inference with this propagation scheme, as an upper bound on the performance of QGNs in a low-memory consumption setting. Similar upper bounds have been used previously to identify the gaps in performance and improve an algorithm (eg. \cite{roig2013active}).

\noindent \textbf{Predicted Composite (PC).} Only pixels predicted by the network to be of the composite class are propagated. This can lead to high sparsity, and also does not require access to ground truth, and is therefore suitable for inference. However, it relies heavily on the network's performance on the composite class, and, in practice, we found this scheme to be unreliable for training from scratch. Gains in performance could be obtained by fine-tuning a model obtained with GTC by using PC, and other variations on the propagation scheme that combine different aspects of All, GTC and PC; but we leave this investigation to future work.

\section{Experiments}
\label{s:exp}
In this section, we demonstrate the benefits of Quadtree Generating Networks on three publicly available semantic segmentation datasets: Cityscapes \cite{cordts2016cityscapes}, SUN-RGBD \cite{song2015sunrgbd} and ADE20K \cite{zhou2018semantic}. We also include ablation studies where we first analyze sparsity of each dataset and then, the memory consumption of our approach compared to typical dilated ResNet based architectures.

\noindent
\textbf{Implementation.} Our QGN encoder and decoder architectures are based on ResNets \cite{he2016deep}. We use the ResNet-50 architecture as our encoder unless otherwise specified. For our decoder, we use a transposed ResNet, similar to the network proposed in \cite{jiang2018rednet}. To implement sparse convolutions in the decoder, we use the \textit{SparseConvNet} framework for PyTorch, introduced with submanifold sparse convolutional networks \cite{graham2018semantic}. Please refer to the supplementary material for more details regarding the experimental setup and decoder architecture.

\begin{table}[t]
	\centering
	\caption{\textbf{Cityscapes:} Pixel accuracy (\%) and mean IoU (\%) of the proposed approach with different propagation schemes and loss weighting mechanisms. Adaptive loss weighting, and fixed weighting with $\gamma=1$, give the best results.}
	\small
	\label{t:prop_schemes}
	\setlength{\tabcolsep}{2pt}
	\begin{tabular}{cc|cc|cc|cc}
	\hline
	\multicolumn{2}{c|}{}&\multicolumn{6}{c}{\small \textbf{Eval Mode}}\\
		\cline{3-8}
		\textbf{Train Mode} & \textbf{Loss} & \multicolumn{2}{c|}{ \textbf{\small{All}}}&\multicolumn{2}{c|}{\small \textbf{GTC}}&\multicolumn{2}{c}{\small \textbf{PC}}\\
		\cline{3-8}
		& & \textbf{Acc} & \textbf{mIoU} & \textbf{Acc} & \textbf{mIoU} & \textbf{Acc} & \textbf{mIoU}\\\hline
		\multirow{4}{*}{All}& $\gamma=0.75$ & 96.25 & 77.49 & 95.84 & 72.17 & 95.38 & 71.13 \\
		& $\gamma=1$ & 96.24 & 77.95 & 95.94 & 72.97 & {95.42} & {71.53}\\
		& $\gamma=1.25$ & 96.17 & 77.25 & {95.95} & {73.02} & 95.35 & 71.52 \\
		& Adaptive & \textbf{96.33} & \textbf{78.20} & 95.76 & 72.89 & 95.20 & 71.33 \\
		\hline
		\multirow{4}{*}{GTC} & $\gamma=0.75$ & 15.88 & 13.58 & 97.76 & 80.13 & 95.16 & 72.16 \\
		& $\gamma=1$ & 32.59 & 17.25 & 97.69 & 80.67 & \textbf{95.42} & 72.44 \\
		& $\gamma=1.25$ & 19.98 & 14.39 & 97.66 & 80.31 & 95.41 & 72.56 \\
		& Adaptive & 31.00 & 18.25 & \textbf{97.81} & \textbf{81.50} & 95.33 & \textbf{73.00} \\
		\hline
	\end{tabular}
	\vspace{-0.5cm}
\end{table}

\subsection{Results on Cityscapes}

Cityscapes \cite{cordts2016cityscapes} is a street scene segmentation dataset, consisting of $2048 \times 1024$ fine pixel-annotated images collected from streets in various European cities. There are 2,975 images in the training set and 500 images for validation. The segmentation task is over 19 commonly observed classes in these street scenes, such as road, car, pedestrian, traffic sign, etc. The high-resolution images in this dataset make existing segmentation networks consume large amounts of memory. QGNs are an ideal solution to address this challenge.

In our first experiment, we evaluate the three propagation schemes (All, GTC and PC as detailed in Section \ref{ss:prop}), when QGNs are trained using All and GTC. While GTC evaluation is not practically applicable, the gap in performance between GTC and PC evaluation helps us to better understand the strengths and weaknesses of the proposed algorithm. We also compare the two loss weighting approaches from Eq. \ref{e:fix_loss} and Eq. \ref{e:adapt_loss} in this experiment. Our results are summarized in Table \ref{t:prop_schemes}.

As shown, when training with the All propagation scheme (that has higher memory consumption), best results are observed for adaptive loss weighting, when the network is also evaluated with the All propagation scheme, achieving 78.20 mIoU. When evaluated with GTC or PC, the different loss weighting approaches do not significantly impact performance, and the network obtains 73.02 and 71.53 mIoU respectively. Importantly, this shows that the same networks that obtain 78.20 mIoU when evaluated with All, can be adaptively modified at inference time with a different propagation scheme such as PC, and still achieve competitive mIoU.

In comparison, networks trained with GTC perform poorly in the All evaluation mode. This is likely due to the fact that the training distribution observed by the deeper layers in the GTC decoder is very different from the training distribution for All, and the final prediction layer is not suitable for assigning labels to all the pixels in an image. When evaluating with the GTC propagation scheme for this model, we observe the best results for adaptive loss weighting. The obtained 81.50 mIoU is an upper bound on performance with QGNs in the low-memory consumption regime. Interestingly, this performance exceeds that of state-of-the-art architectures such as DeepLabv3 and CCNet, which rely on deeper networks and additional context modules. Finally, while evaluating with PC, both $\gamma=1$ and adaptive loss weighting perform well. The gap in performance between GTC and PC evaluation helps us isolate the impact of the quadtree 'structure prediction' and final 'label assignment' tasks performed by the QGN on performance. Based on the observed gap (8.5 points), we see that 'structure prediction' is the main challenge for improving our approach.

\begin{table}[!t]
	\begin{center}
		\small
		\caption{\textbf{Cityscapes:} Activation memory consumption (GB), FLOPs ($\times10^{12}$), and mean IoU (\%) of the proposed approach in comparison to state-of-the-art dilation-based methods on the validation set, for input resolution $2048\times1024$. QGN results are with adaptive loss weighting. Our approach significantly outperforms dilated networks with similar memory consumption, and provides competitive results to state-of-the-art models that are computationally more intensive.}
		\label{t:cityscapesSOTA}
		\setlength{\tabcolsep}{3pt}
		\begin{tabular}{cc|cc|c}
		\hline
		\textbf{Method} & \textbf{Model} & \textbf{Memory} & \textbf{TFLOPs} & \textbf{mIoU}\\
		\hline
		\multirow{4}{*}{Dilation} & DRN-C-42 \cite{yu2016dilated} & 3.77 & 1.07 & 70.9 \\
		& DRN-D-105 \cite{yu2016dilated} & 15.15 & 1.91 & 75.6 \\
		& DeepLabv3 \cite{chen2017rethinking} & 14.27* & 1.97* & 79.3 \\
		& CCNet \cite{huang2018ccnet} & 14.33* & 1.55* & 79.8 \\
		\hline
		\multirow{2}{*}{QGN (Ours)} & Train-All-Eval-All & 5.85 & 0.48 & 78.2 \\
		& Train-GTC-Eval-PC & 3.66 & 0.25 & 73.0 \\
		\hline
	\end{tabular}
	\footnotesize{\ \ * indicates an estimate based on architecture details}
	\end{center}
	\vspace{-1.0cm}
\end{table}

\noindent
\textbf{Comparison to dilated convolutions.} We focus now on comparing our best results with PC evaluation to existing semantic segmentation networks that use dilated convolutions in the encoder. Table~\ref{t:cityscapesSOTA} summarizes our results in terms of memory consumption, FLOPs and mIoU. As shown, QGNs with Train-GTC-Eval-PC provide a 3\% relative mIoU improvement over the DRN-C-42 network, which consumes a similar amount of memory. Further, they only have a 3\% less relative mIoU compared to the DRN-D-105 network, which consumes $4\times$ more memory and $6\times$ more compute. DeepLabv3 and CCNet provide slightly better results than our proposal. However, they require deeper backbone networks, specialized context modules, and at least $2.5\times$ more memory and $3\times$ more compute than QGNs.

From these results, we can conclude that adaptive loss weighting, and fixed weighting with $\gamma=1$, give the best performance. Further, training with GTC leads to maximum performance when evaluating with PC, but training with All gives networks that can be adaptively modified at inference time with competitive results when evaluated with all 3 propagation schemes. Of special interest is the low memory requirement of our approach (see Table~\ref{t:cityscapesSOTA}): contrary to methods such as DeepLabv3 or CCNet which need to process full resolution images in parts on a standard GPU with 12GB of memory, our proposal can process 2 or 3 frames in parallel on the same GPU at full resolution. As a result, our architecture could be used to process consecutive frames in batches allowing for improved throughput, or perform multi-scale fusion during inference to improve performance.

\begin{table}[t]
	\centering
	\caption{\textbf{SUN-RGBD:} Pixel accuracy (\%) and mean IoU (\%) of the proposed approach on the 13-way and 37-way validation sets. All methods use $\gamma=1$. Relative performance of the 3 propagation schemes is similar across both datasets.}
	\small
	\label{t:sunrgbd}
	\setlength{\tabcolsep}{4pt}
	\begin{tabular}{c|cc|cc|cc}
		\hline
	    &\multicolumn{6}{c}{\small \textbf{Eval Mode}}\\
		\cline{2-7}
		\textbf{Train Mode} & \multicolumn{2}{c|}{ \textbf{\small{All}}}&\multicolumn{2}{c|}{\small \textbf{GTC}}&\multicolumn{2}{c}{\small \textbf{PC}}\\
		\cline{2-7}
		& \textbf{Acc} & \textbf{mIoU} & \textbf{Acc} & \textbf{mIoU} & \textbf{Acc} & \textbf{mIoU}\\\hline
		\multicolumn{7}{c}{\small \textbf{SUN-RGBD-13}}\\
		\hline
		All & \textbf{85.32} & \textbf{62.70} & 81.92 & 55.76 & 82.59 & 56.65\\
        GTC & 66.91 & 36.03 & \textbf{87.12} & \textbf{64.57} & \textbf{84.08} & \textbf{59.19} \\
		\hline
		\multicolumn{7}{c}{\small \textbf{SUN-RGBD-37}}\\
		\hline
		All & \textbf{80.69} & \textbf{44.11} & 77.26 & 37.88 & 78.05 & 39.13 \\
		GTC & 65.24 & 28.10 & \textbf{82.36} & \textbf{45.36} & \textbf{79.26} & \textbf{41.10} \\
		\hline
	\end{tabular}
	\vspace{-0.5cm}
\end{table}

\subsection{Results on SUN-RGBD}
The focus of our next experiment is to analyze the impact of increasing the number of classes on different propagation schemes. To this end, we use the {SUN-RGBD} dataset \cite{song2015sunrgbd}, an indoor scene segmentation benchmark that consists of two segmentation tasks (13 and 37 categories) with 5,285 RGB-D training images and 5,050 test images.  We use the same experimental setting as the one used for the first Cityscapes experiment.

Table \ref{t:sunrgbd} summarizes our results. As shown, with respect to the propagation scheme, the trend remains similar to Cityscapes: the GTC upper bound provides the best performance, followed by All, and then PC. On SUN-RGBD-37, our QGN based on a ResNet-50 encoder evaluated using All provides competitive results (44.11 mIoU) to the 45.90 of RefineNet \cite{lin2017refinenet} which builds upon a ResNet-152 backbone, and is the state-of-the-art for this task. Regarding the influence of the number of classes, our results show no significant difference in terms of relative mIoU of the best GTC and PC results with respect to All (1.03 and 094 for SUN-RGBD-13; and 1.03 and 0.93 on SUN-RGBD-37). These results suggest that the number of classes has no significant impact on the performance of our approach.

\begin{table}[t]
	\centering
	\caption{\textbf{ADE20k:} State-of-the-art performance comparisons in terms of pixel accuracy (\%) and mean IoU (\%) on the validation set. All QGN results are with the Train-All-Eval-All models, and $\gamma=1$. All Dilation based numbers are obtained from the respective papers. With PSPNet, QGNs outperform Dilation, and provide competitive results to other recent context modules in literature.}
	\small
	\label{t:ade20k}
	\setlength{\tabcolsep}{4pt}
	\begin{tabular}{c|c|c|cc}
		\hline
		\textbf{Context} & \textbf{Encoder} & \textbf{Method} & \textbf{Acc} & \textbf{mIoU} \\\hline
		\hline
		\multirow{4}{*}{PSPNet \cite{zhao2017pspnet}} & \multirow{2}{*}{ResNet-50} & Dilation & 80.76 & 42.78\\
		& & QGN (Ours) & 80.43 & 41.42\\
		\cline{2-5}
		& \multirow{2}{*}{ResNet-101} & Dilation & 81.39 & 43.29 \\
		& & QGN (Ours) & 81.67 & 43.91 \\
		\hline
		\hline
		SAC \cite{zhang2017scale} & \multirow{4}{*}{ResNet-101} & \multirow{4}{*}{Dilation} & 81.86 & 44.30\\
		EncNet \cite{zhang2018context} & & & 81.69 & 44.65\\
		PSANet \cite{zhao2018psanet} & & & 81.51 & 43.77\\
		CCNet \cite{huang2018ccnet} & & & - & 45.22 \\
		\hline
	\end{tabular}
	\vspace{-0.1cm}
\end{table}

\begin{table}[t]
	\centering
	\caption{Percentage of pixels in the three datasets (at original resolutions) assigned to each level in a 6-level quadtree representation. The final column indicates the average ratio of memory occupied by the efficient quadtree representation in comparison to dense pixel grid segmentation masks for that dataset (in \%). We observe that a large fraction of the pixels belong to larger cells, with less than 3\% at the leaf ($Q_0$) nodes, leading to compression ratios of 95\% or more. See supplemental for a further visual comparison of quadtree representations and dense pixel grids.}
	\small
	\setlength{\tabcolsep}{2pt}
	\begin{tabular}{cc|cccccc|c}
		\hline
		\textbf{Dataset} & \textbf{Split} & $Q_5$ & $Q_4$ & $Q_3$ & $Q_2$ & $Q_1$ & $Q_0$ & \textbf{Ratio}\\
		\hline
		\multirow{2}{*}{Cityscapes} & train & 66.34 & 14.21 & 9.18 & 5.52 & 3.02 & 1.70 & 3.07 \% \\
		& val & 65.12 & 14.44 & 9.53 & 5.85 & 3.22 & 1.81 & 3.25 \% \\
		\hline
		\multirow{4}{*}{SUN-RGBD} & train-13 & 56.26 & 18.22 & 11.78 & 7.09 & 4.20 & 2.43 & 4.24 \% \\
		& val-13 & 55.57 & 18.50 & 11.98 & 7.22 & 4.27 & 2.46 & 4.29 \% \\
		& train-37 & 56.11 & 18.27 & 11.82 & 7.12 & 4.22 & 2.44 & 4.25 \% \\
		& val-37 & 55.40 & 18.54 & 12.03 & 7.25 & 4.29 & 2.47 & 4.31 \% \\
		\hline
		\multirow{2}{*}{ADE20k} & train & 47.48 & 21.40 & 14.44 & 8.68 & 5.05 & 2.93 & 5.09 \% \\
		& val & 47.25 & 21.44 & 14.49 & 8.74 & 5.10 & 2.96 & 5.14 \% \\
		\hline
	\end{tabular}
	\label{t:datasets}
	\vspace{-0.5cm}
\end{table}

\begin{table*}[t]
	\centering
	\caption{Average forward pass activation memory consumption with different model architectures (in GB) for segmentation of an image from the validation datasets used in our experiments. R50 denotes ResNet-50 and R101 denotes ResNet101. All and GTC correspond to the propagation schemes detailed in Section \ref{s:QGN}. Dil denotes dilation in the encoder, and the corresponding 'Decoder Only' memory refers to a single $1\times1$ convolution as a decoder. The architectures with the \textbf{lowest} and \textit{highest} memory consumption for each backbone are marked in \textbf{bold} and \textit{italic} respectively. QGNs consume significantly less memory than dilation based networks.}
	\small
	\setlength{\tabcolsep}{9pt}
	\begin{tabular}{c|ccc|ccc|ccc}
		\hline
		\textbf{Architecture} &\multicolumn{3}{c|}{\textbf{Decoder Only}}&\multicolumn{3}{c|}{\textbf{Encoder-Decoder}}&\multicolumn{3}{c}{\textbf{Encoder-Decoder}}\\
		\hline
		\textbf{Dataset} & \textbf{Dil} & \textbf{All} & \textbf{GTC}  & \textbf{R50-Dil} & \textbf{R50-All} & \textbf{R50-GTC} & \textbf{R101-Dil} & \textbf{R101-All} & \textbf{R101-GTC}\\
		\hline
		Cityscapes & 0.15 & 2.28 & 0.09 & \textit{7.52} & 5.85 & \textbf{3.66} & \textit{13.89} & 7.44 & \textbf{5.26} \\
		SUN-RGBD-13 & 0.01 & 0.27 & 0.02 & \textit{0.98} & 0.71 & \textbf{0.46} & \textit{1.72} & 0.91 & \textbf{0.66} \\
		SUN-RGBD-37 & 0.04 & 0.31 & 0.03 & \textit{1.01} & 0.75 & \textbf{0.47} & \textit{1.75} & 0.95 & \textbf{0.67} \\
        ADE20k & 0.17 & 0.53 & 0.08 & \textit{1.31} & 1.05 & \textbf{0.60} & \textit{2.18} & 1.29 & \textbf{0.84} \\
		\hline
	\end{tabular}
	\label{t:mem_foot}
	\vspace{-0.3cm}
\end{table*}

\subsection{Results on ADE20k}

{ADE20k} \cite{zhou2018semantic,zhou2017scene} is a recent scene parsing benchmark with a large number of classes (150 total classes, with 35 stuff classes, i.e., road, sky, wall... which occupy 61\% of the pixels in the dataset, and 115 thing classes, i.e., table, person, car... which occupy 32\% of the pixels). The dataset consists of natural images of a variety of indoor and outdoor scenes, with 20,210 training images and 2,000 validation images. The large proportion of thing classes makes ADE20k a challenging task that requires a lot of contextual reasoning. In order to counter these challenges, existing results involve the use of dilated convolutions, in combination with a specialized module after the encoder that improves contextual reasoning \cite{zhao2017pspnet,zhang2017scale,zhang2018context,zhao2018psanet,huang2018ccnet}.

For our experiments on this task, we use a modified version of QGNs, inserting a PSPNet \cite{zhao2017pspnet} module in between our encoder and decoder. This allows us to check the compatibility of QGNs with specialized context modules. We compare our approach to state-of-the-art results in terms of pixel accuracy and mIoU with both ResNet-50 and ResNet-101 backbones.

Our results are shown in Table \ref{t:ade20k}. With PSPNet, QGNs outperform dilation on ResNet-101, showing the potential of QGNs as a generic approach that can be paired with existing context modules in literature. The achieved 43.91 mIoU is also competitive to the state-of-the-art obtained with more recent context modules. Our results indicate that there may be further gains by using QGNs as a drop-in replacement for dilation in these newer segmentation approaches.

\subsection{Memory footprint analysis}

\noindent
\textbf{Dataset sparsity.} To evaluate the efficiency of quadtree representations, we convert the provided ground truth segmentation masks in each dataset into 6-level quadtrees and analyze the sparsity of the representation in Table \ref{t:datasets}. We observe that there is significant redundancy across all datasets, with less than 3\% of the pixels in the dataset being assigned to the leaf nodes of the quadtree. The proposed representation eventually occupies orders of magnitude less memory than the typical dense pixel grid. The compression ratio is maximum (around $97\%$, or $33\times$) for higher-resolution images with few classes (Cityscapes) and minimum (around $95\%$, or $20\times$) for lower-resolutions with more classes (ADE20k). Further analysis of the dataset sparsity, in terms of visualizations of the quadtree representation for validation images from the Cityscapes and ADE20k datasets, is provided in the supplemental material.

\noindent
\textbf{Activation memory footprint.} The actual reduction in memory footprint achieved through QGNs depends on a number of factors besides the dataset sparsity, including the input resolution, network architecture and propagation scheme. It is important to note that the reduction in memory footprint with our approach comes only with respect to the feature activations (which form a majority of the occupied memory during training). The memory consumption of the model parameters remains the same for different propagation schemes.

To evaluate the actual memory reduction, we compare the forward pass memory consumption of QGNs with the All and GTC propagation schemes, using ResNet-50 and ResNet101 backbones, to networks based on dilated convolutions. When using dilated convolutions, dilate the last 2 residual blocks of the encoder (Block3 and Block4) so that the feature maps produced are 8 times smaller than the input resolution along each dimension. We remove the QGN decoder, and use a single $1\times1$ convolutional layer as a decoder with these dilated encoders to obtain the logits. We use bilinear interpolation on the logits to recover a prediction at the same resolution as the input. The QGN decoder has a parameter overhead of around 0.12GB as opposed to a single $1\times1$ convolutional layer, but this overhead is negligible compared to the memory gains for QGNs through the activations, as summarized in Table \ref{t:mem_foot}.

For the same backbone network, a QGN encoder-decoder is more memory efficient than a dilated network, even when operating with the All propagation scheme (R50-All and R101-All). Furthermore, when using the GTC propagation scheme, the decoder activations consume significantly less memory ($<$0.1GB), sometimes less than the memory consumed by a single $1\times1$ convolution used as a decoder with dilation. Overall, our GTC approach consumes up to $2.6\times$ less memory than a dilation based network using the same backbone architecture, while maintaining similar performance (see Tables \ref{t:cityscapesSOTA} and \ref{t:ade20k}).

While QGNs are designed to reduce decoder memory, a key advantage is that the QGN decoder \textit{enables us to use a stride 32 backbone without losing performance}. As a result, our decoder is implicitly responsible for the memory reductions observed in the encoder. Existing methods perform poorly when the encoder is replaced with a higher stride network. For example, when compared to a DeepLabv3 decoder + ResNet-50 stride 32 encoder (from \cite{chen2017rethinking}), which only obtains 70.0 mIoU on Cityscapes, our method achieves a 4.3\% better relative mIoU of 73.0 (see Table \ref{t:cityscapesSOTA}).

\section{Conclusion}

In this paper, we presented Quadtree Generating Networks (QGNs) for semantic segmentation. The key idea is to make predictions as a hierarchical quadtree instead of a dense rectangular grid, using a sparse convolutional decoder. Our results on three public benchmarks demonstrate that QGNs can process high-resolution images and outperform those approaches that have similar memory consumption. In addition, our results are competitive when compared to state-of-the-art methods while consuming $2\times$ to $4\times$ less memory. Moreover, our approach is flexible and can be adapted at inference time to trade-off between performance and memory consumption.

\subsection*{Acknowledgements}
We would like to thank Yash Patel for several insightful comments and discussions.

{\small
\bibliographystyle{ieee}
\bibliography{egbib}
}

\onecolumn
\appendix
\section*{\LARGE Supplementary Material}

\setcounter{section}{0}
\section{Implementation Details}
In this section, we provide more information about the network architectures and hyper-parameters used in the experiments from Section \ref{s:exp}.

\subsection{Network Architecture}

Our QGN encoder and decoder architectures are based on ResNets \cite{he2016deep}. Similar to existing semantic segmentation papers, we replace the first $7\times7$ convolution of the original ResNet with three $3\times3$ convolutions \cite{zhao2017pspnet,chen2018encoder,xiao2018upernet}. We use the ResNet-50 architecture as our encoder unless otherwise specified. For our decoder, we use a transposed ResNet, similar to the network proposed in \cite{jiang2018rednet}. Specifically, this decoder is a residual network with 38 layers. The exact channel and layer counts for the encoder and decoder are summarized in Table \ref{t:arch}.

\begin{table}[h!]
    \caption{Encoder and Decoder configurations for QGNs, based on the architecture proposed in \cite{jiang2018rednet}. Each 'Block' subsamples or upsamples the image by a factor of 2.}
    \centering
    \small
    \setlength{\tabcolsep}{5pt}
    \begin{tabular}{cccccccc}
        \hline
        \multirow{2}{*}{Block} &\multicolumn{3}{c}{Encoder} &\multirow{2}{*}{Block} &\multicolumn{3}{c}{Decoder} \\
        \cline{2-4} \cline{6-8}
        &In     &Out      &\# Unit & & In &Out &\# Unit \\
        \hline
        Block0   &3       &64       &-   &Trans4    &512     &256   &6\\
        Block1  &64      &256      &3   &Trans3    &256     &128   &4\\
        Block2  &256     &512      &4   &Trans2    &128     &64    &3\\
        Block3  &512     &1024     &6   &Trans1    &64      &64    &3\\
        Block4  &1024    &2048     &3   &Trans0    &64      &64    &3\\
        \hline
    \end{tabular}
    \label{t:arch}
\end{table}

\noindent
\textbf{Sparse convolutions.} To implement sparse convolutions in the decoder, we use the \textit{SparseConvNet} framework, introduced with submanifold sparse convolutional networks \cite{graham2018semantic}. We also implement a sanity check that uses dense convolutions with multiplicative masks of zeros to emulate sparse convolutions. We found that the wall-clock time for training with sparse vs. dense convolutions is similar (up to  +10\% training time depending on dataset sparsity). More optimized implementations of sparse convolutions (eg. \cite{ren2018sbnet}) could translate the large gains in memory efficiency to reductions in wall-clock training time.

\subsection{Optimization}

\noindent
\textbf{Input resolution.} We take advantage of our efficient architecture by training QGNs with full resolution input images (without cropping) on all datasets. Within the ADE20k and SUN-RGBD datasets, resolution and aspect ratios vary widely ($561\times427$ to $730\times530$, with many others). We found that zero padding to the nearest multiple of 32 pixels (maximum downsampling ratio of the encoder) during training effectively deals with this problem. During inference, we use bilinear interpolation to rescale the image dimensions to the nearest multiple of 32, make the predictions at each pixel, and rescale the resulting segmentation map to the original input dimensions with nearest neighbor interpolation.

\noindent
\textbf{Learning rate schedule.} We optimize our networks with Stochastic Gradient Descent (SGD), with an initial learning rate $\alpha_0$ of $0.02$, which is decreased based on a polynomial decay function $\alpha = \alpha_0 (1-\frac{i}{i_{max}})^{\rho}$. We use a decay factor $\rho$ of $0.9$, training for a total of $i_{max}=100,000$ iterations. Random scaling and horizontal flips are applied for data augmentation. For all our experiments, we maintain this set of hyper-parameters, without specific tuning to each task.

\noindent
\textbf{Class weighting.} To counter the effect of class imbalance, we double the weight of the loss for the set of classes whose IoU is below the median IoU on a test subset. Specifically, we evaluate the performance of the model every $5,000$ training iterations on a subset of $100$ images set aside for validation, and use these results to set the class weights.

\section{Dataset Sparsity Visualization}
In this section, we visualize the quadtree representations of the segmentation maps obtained using the procedure detailed in Section \ref{s:quadtree}. Specifically, we are interesting in visualizing each level of the ground truth quadtree representation. To this end, we apply the merge operator from Eq. \ref{e:merge_op} to validation images from the Cityscapes and ADE20k datasets. The visualizations we obtain are presented in Table \ref{t:city_vis} and Table \ref{t:ade_vis} respectively.

We observe that a majority of the image pixels have their ground truth encoded in $Q_5$, which is highly efficient due to its lower resolution. The intermediate levels coarsely encode semantic boundaries, with the leaf nodes ($Q_0$) encoding the finer details of these boundaries between classes. These observations mirror the statistics observed in Table \ref{t:datasets}, showcasing the huge potential for lossless compression in semantic segmentation using quadtree representations.

\begin{sidewaystable}
\centering
\caption{Visualization of dataset sparsity using examples from the Cityscapes validation set. ${Q}_5, ..., {Q}_0$ are the different quadtree levels. As noted in Table \ref{t:datasets}, a large fraction of the pixels belong to larger cells (such as ${Q}_5$), while the leaf ($Q_0$) nodes are extremely sparse.  The composite class is represented in black (best viewed in color).}
\begin{tabular}[\textheight]{c|ccccccc}
		\hline
		Image & $Q_5$ & $Q_4$ & $Q_3$ \\
        \includegraphics[width=0.19\textwidth]{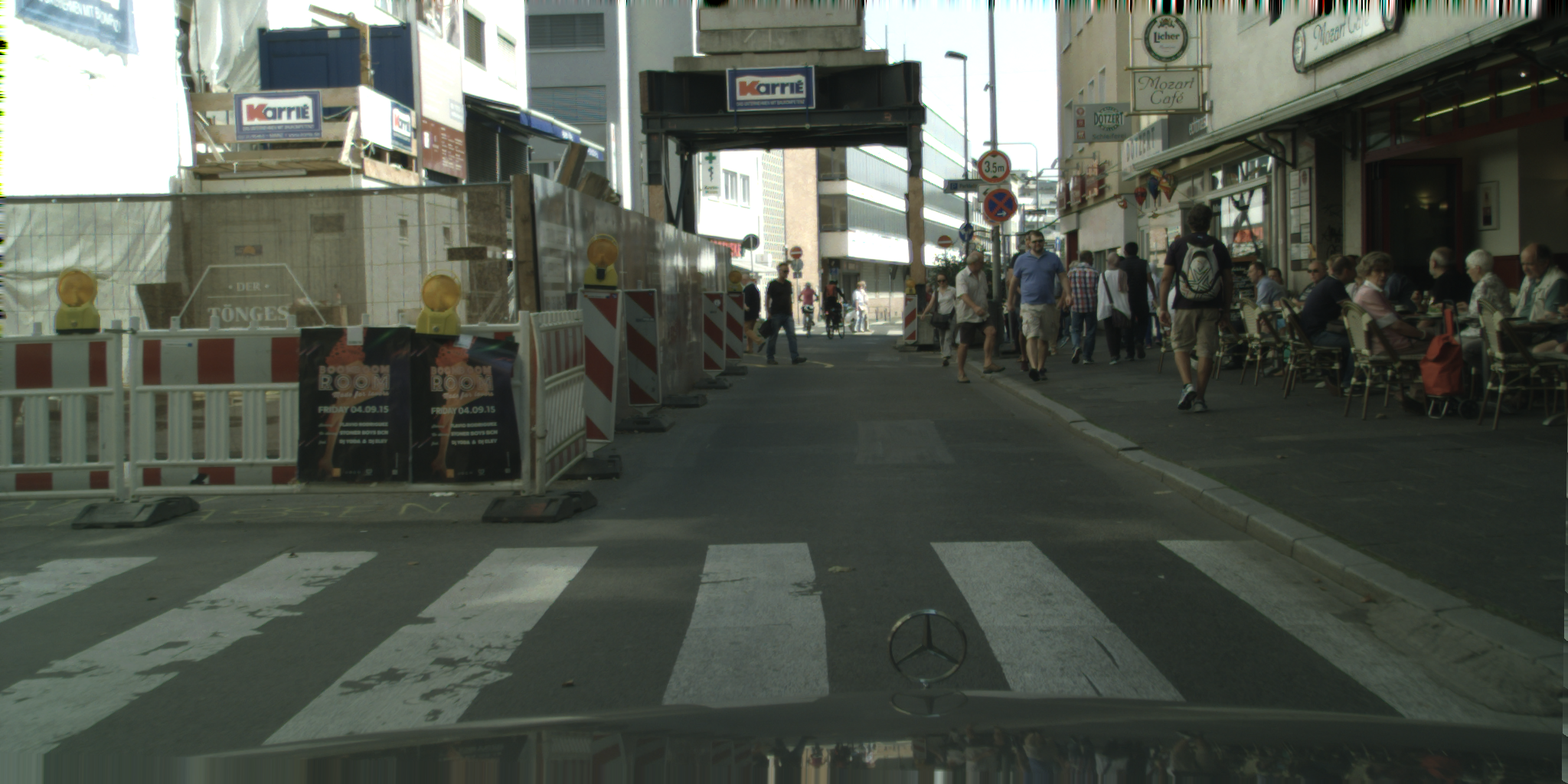} &
        \includegraphics[width=0.19\textwidth]{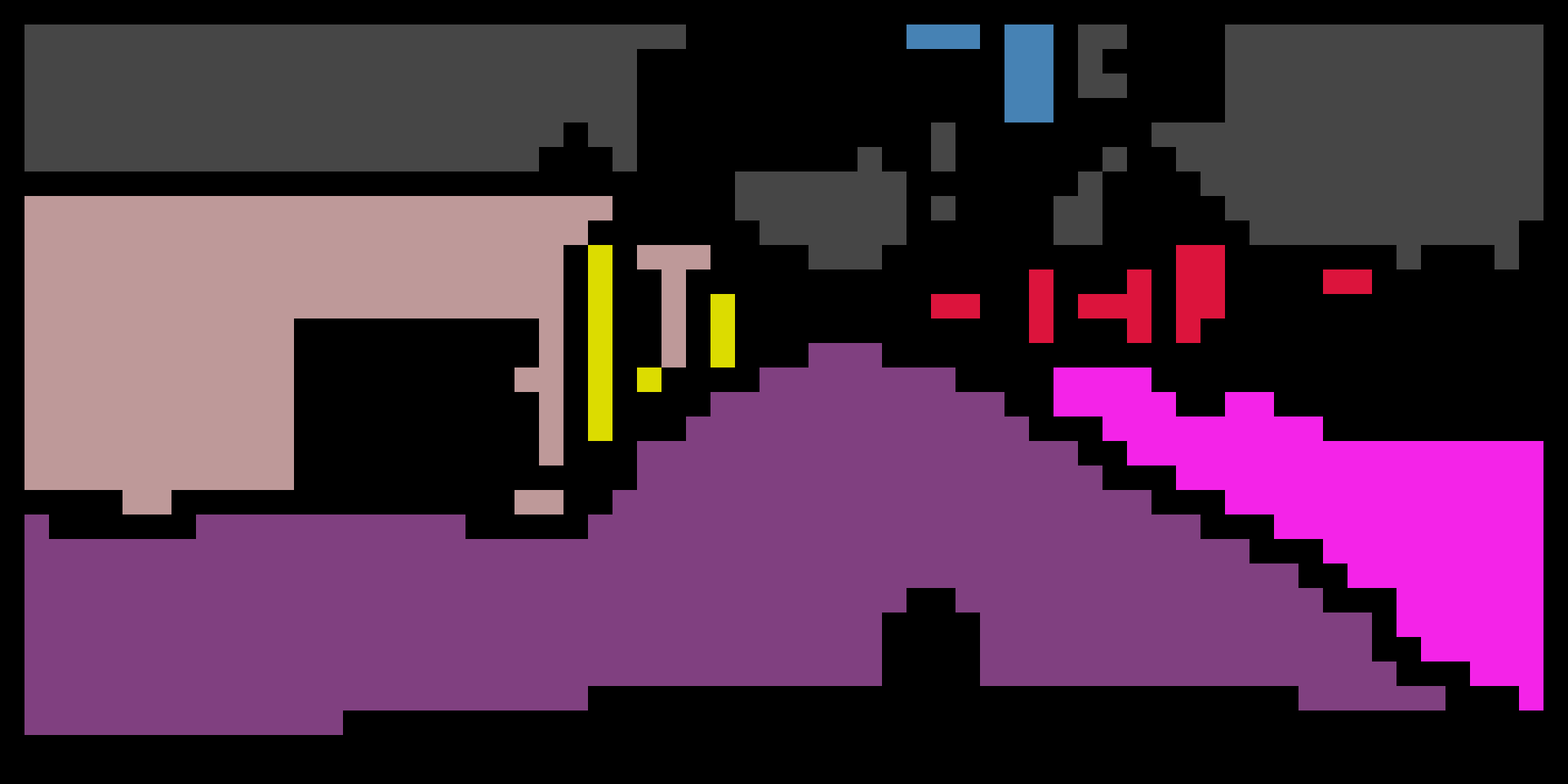} &
        \includegraphics[width=0.19\textwidth]{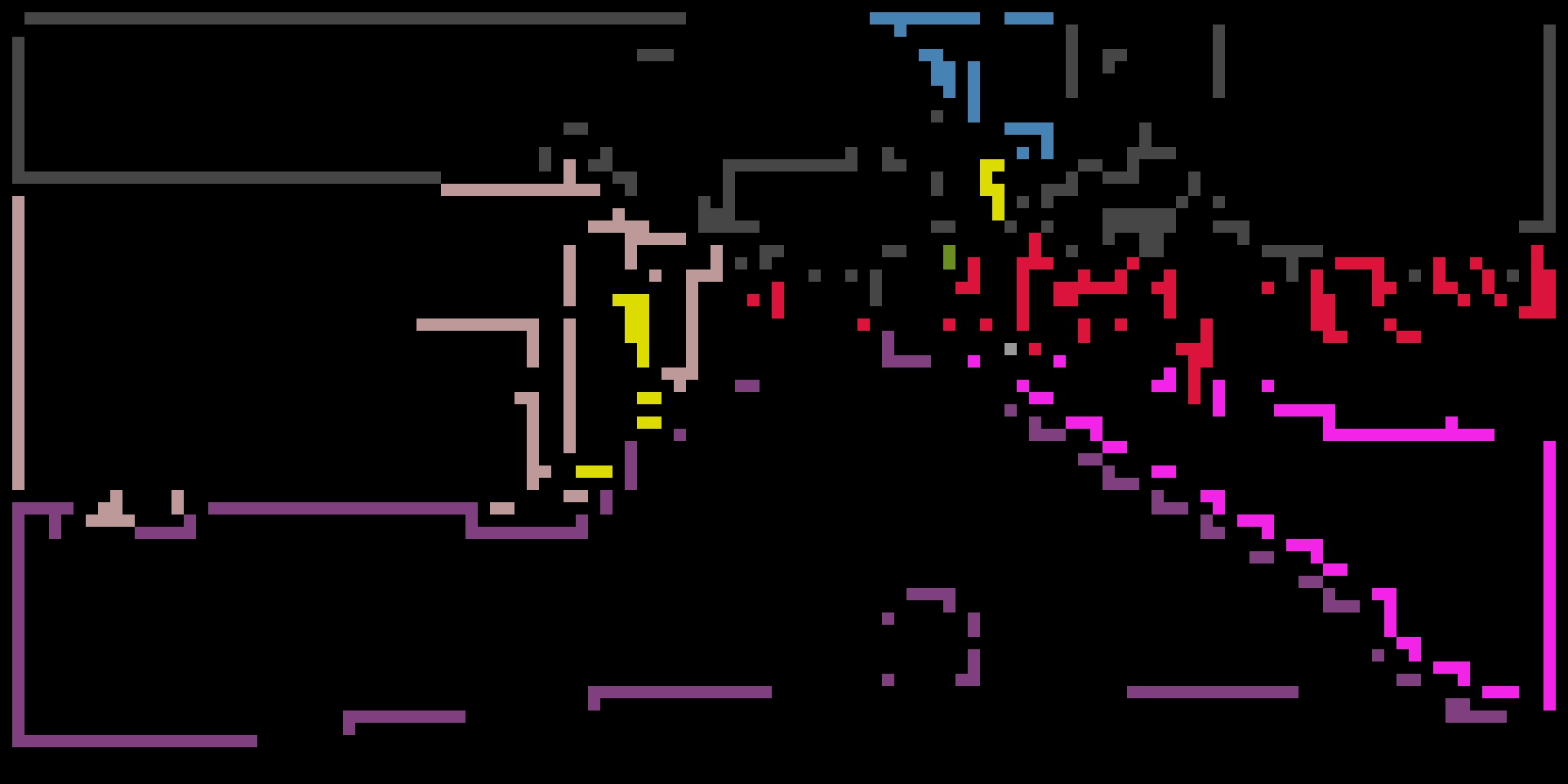} &
        \includegraphics[width=0.19\textwidth]{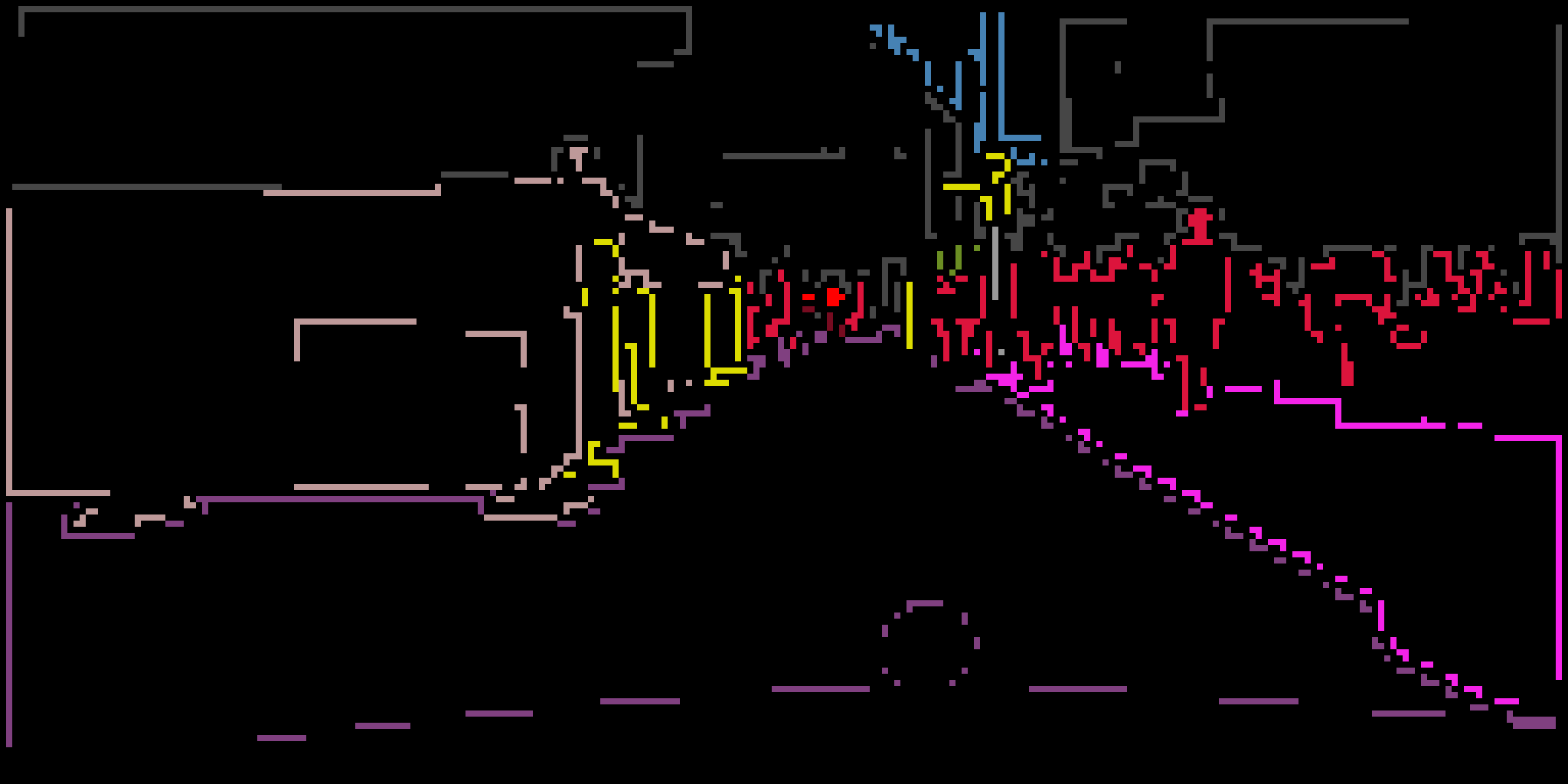} \\
        Segmentation Mask & $Q_2$ & $Q_1$ & $Q_0$\\
        \includegraphics[width=0.19\textwidth]{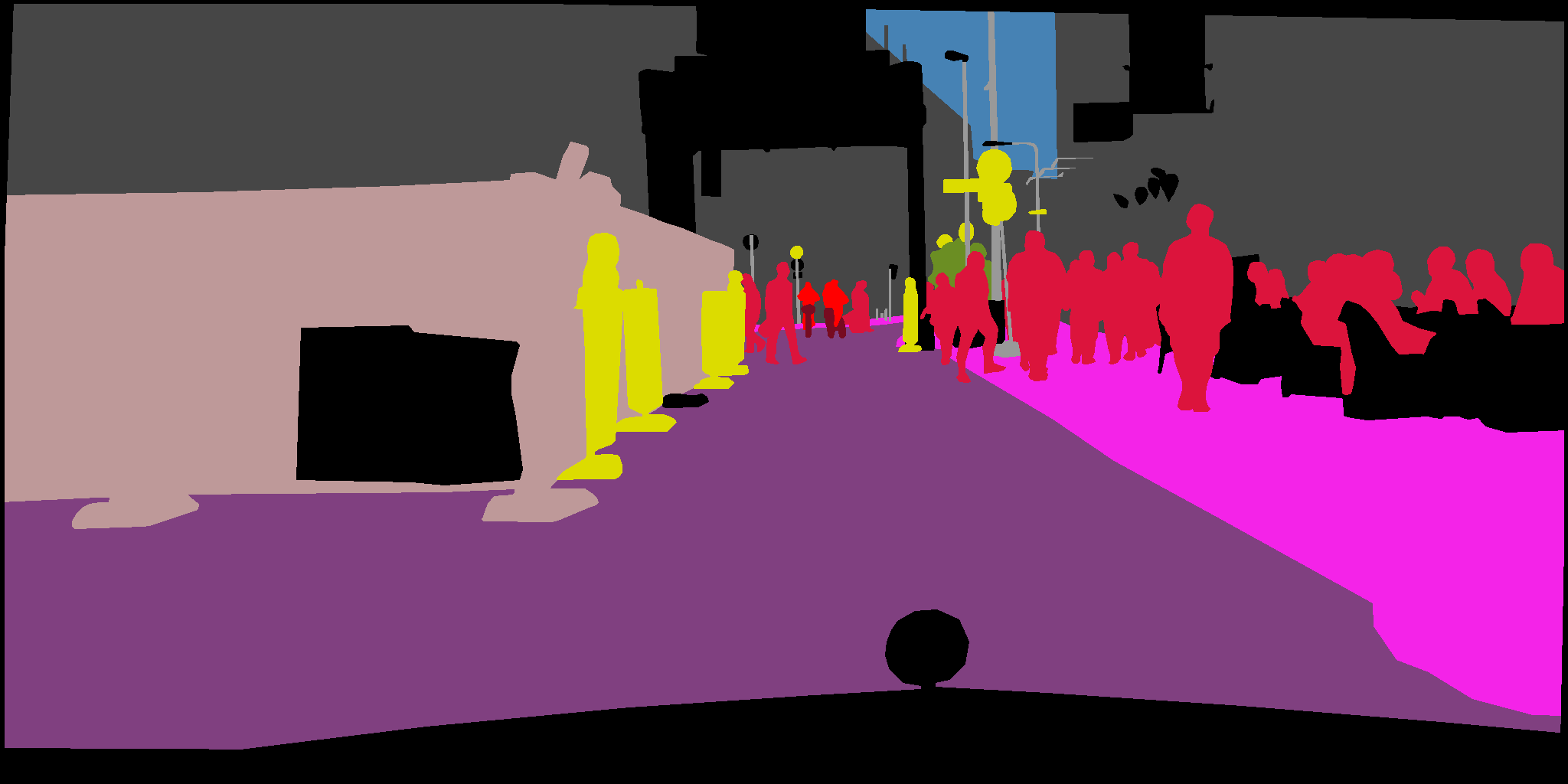} & \includegraphics[width=0.19\textwidth]{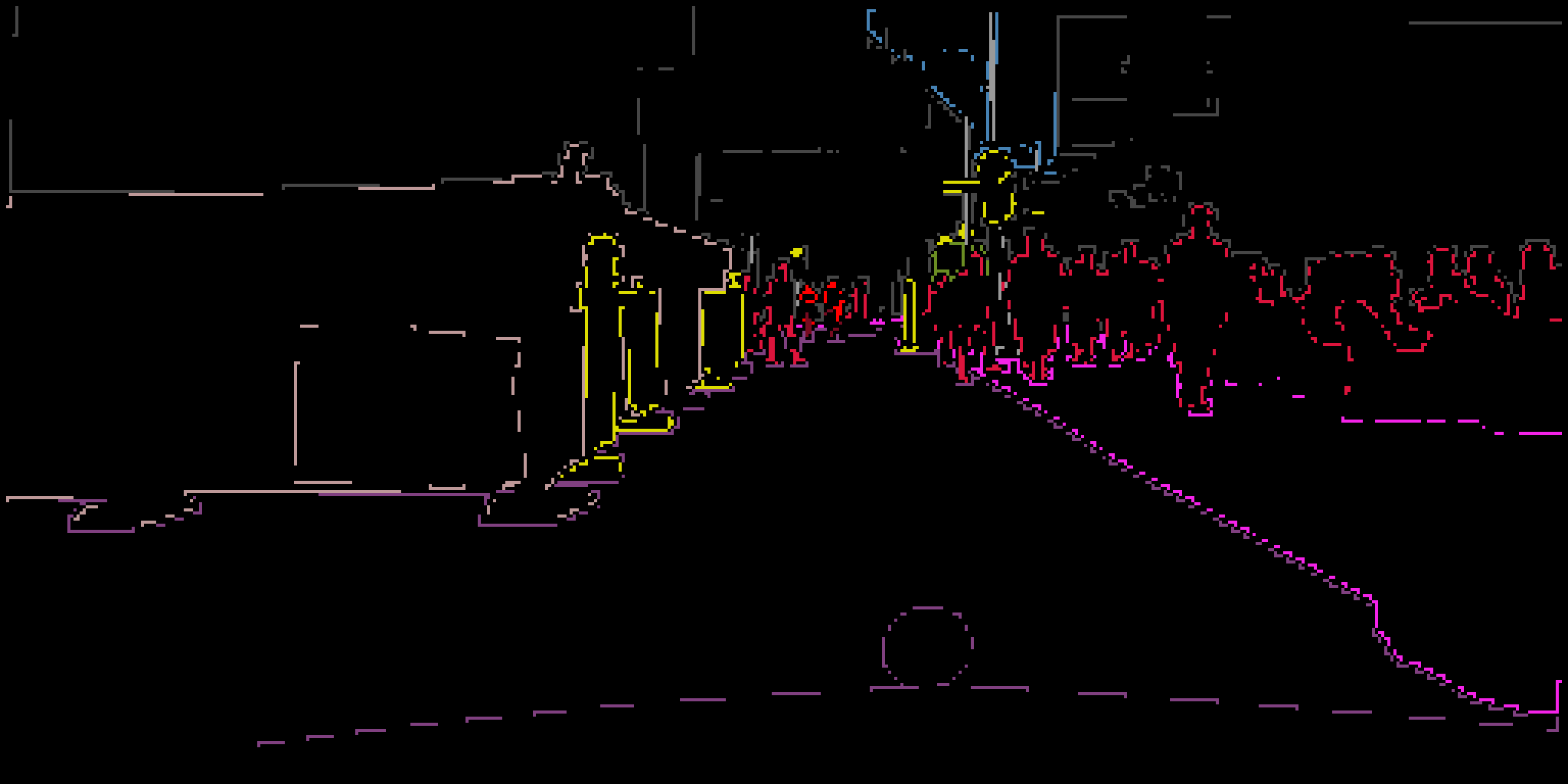} &
        \includegraphics[width=0.19\textwidth]{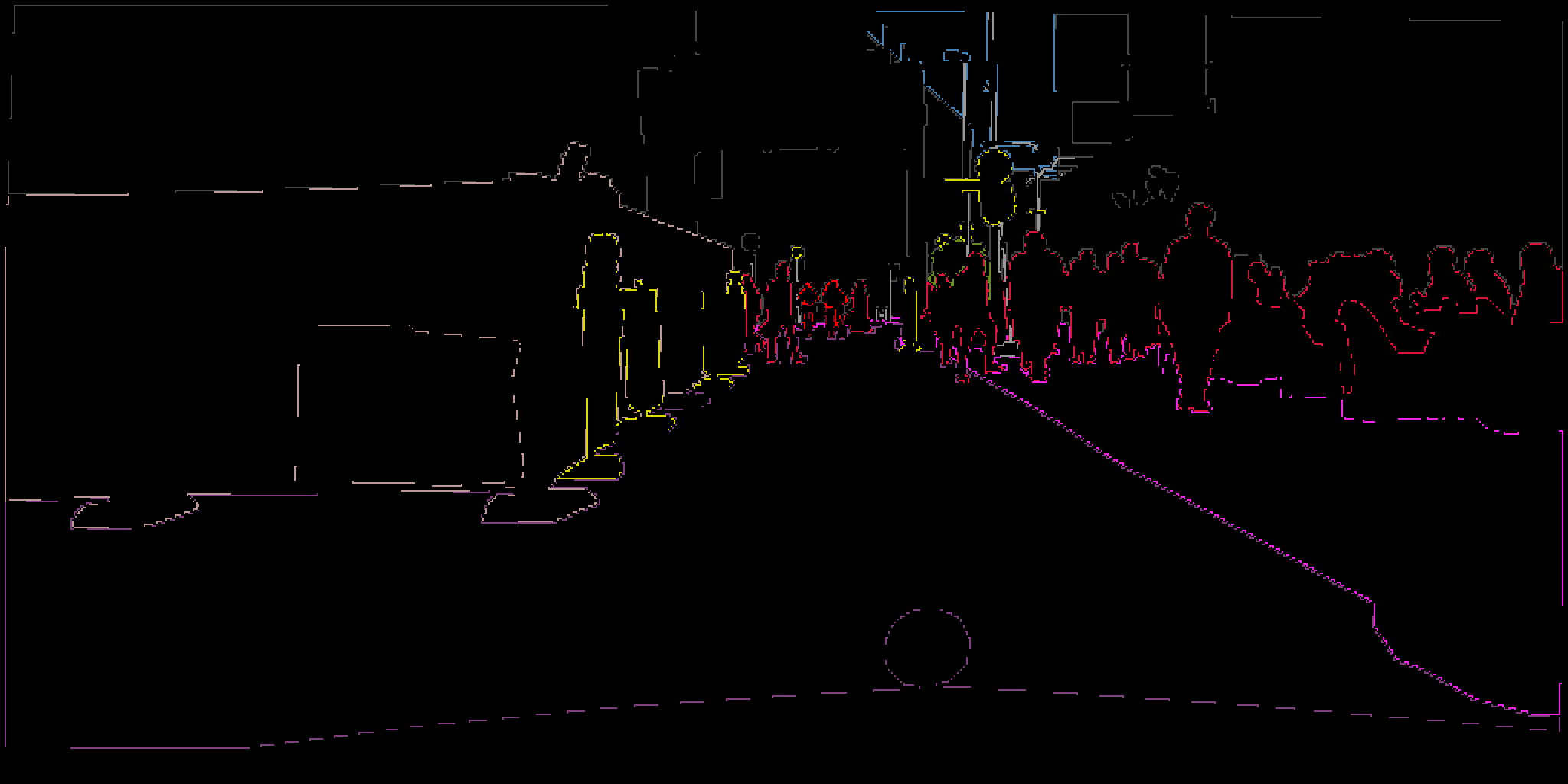} &
        \includegraphics[width=0.19\textwidth]{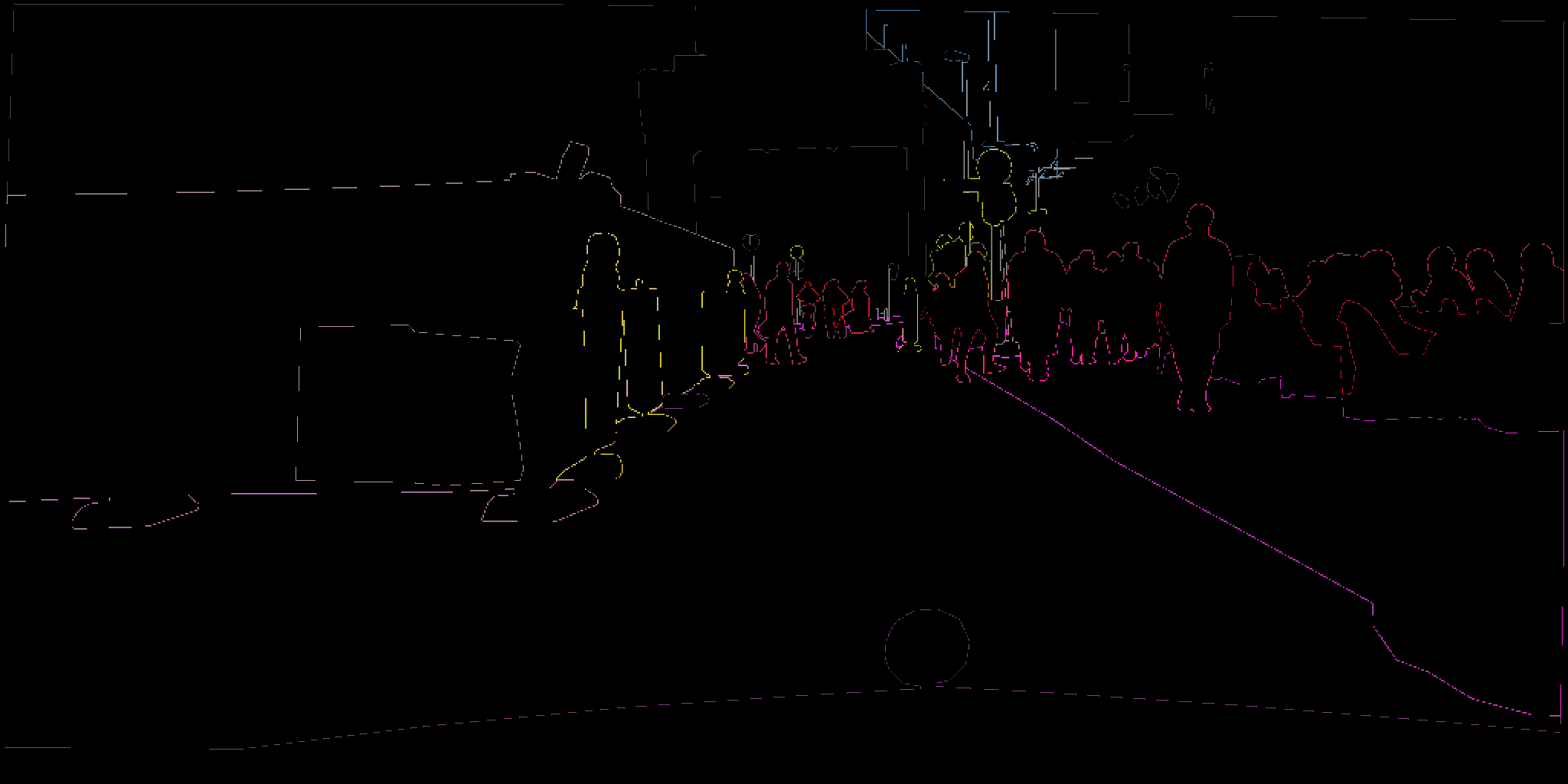} \\
		\hline
		\\
		\hline
		Image & $Q_5$ & $Q_4$ & $Q_3$ \\
		\includegraphics[width=0.19\textwidth]{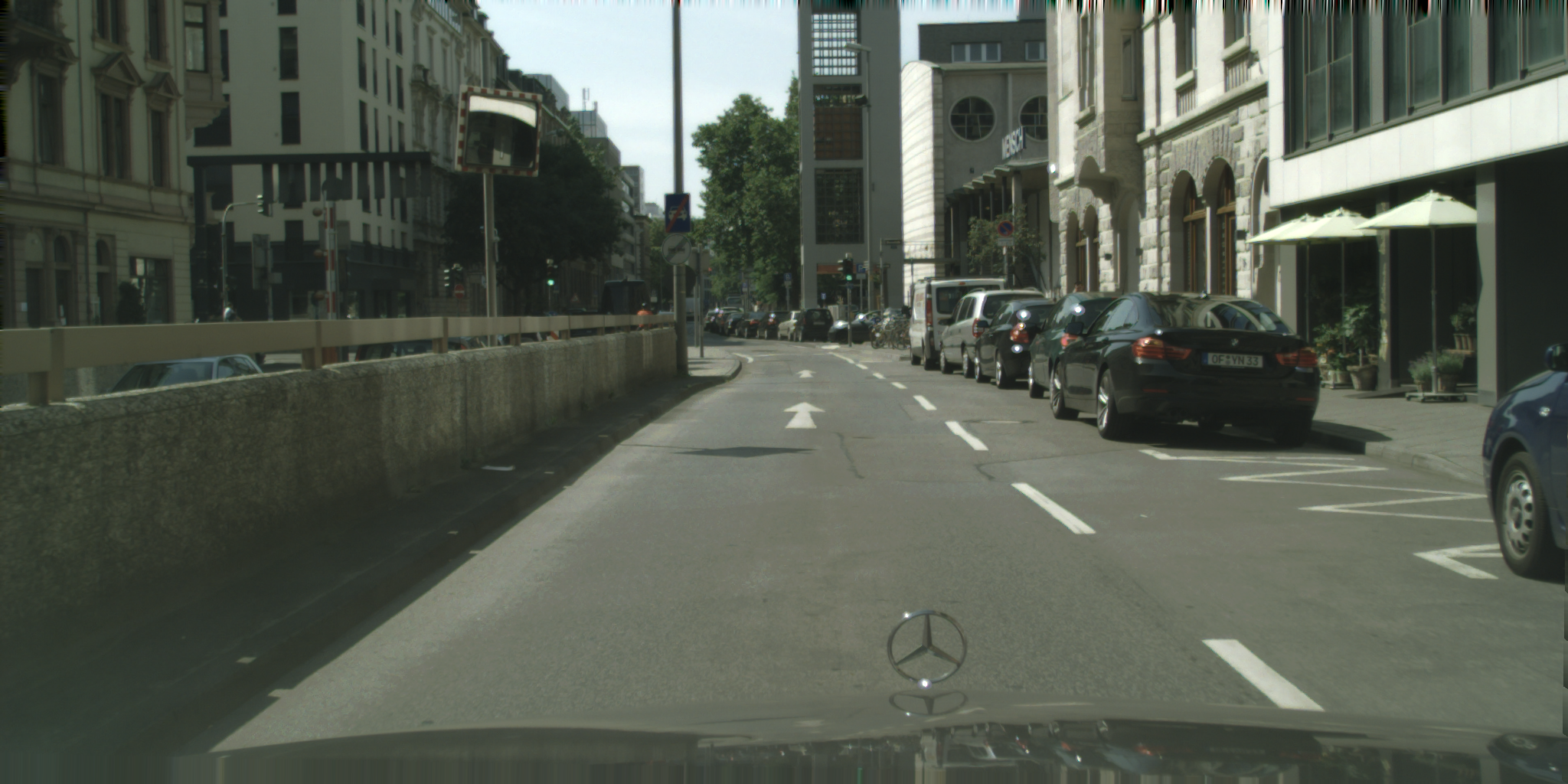} &
		\includegraphics[width=0.19\textwidth]{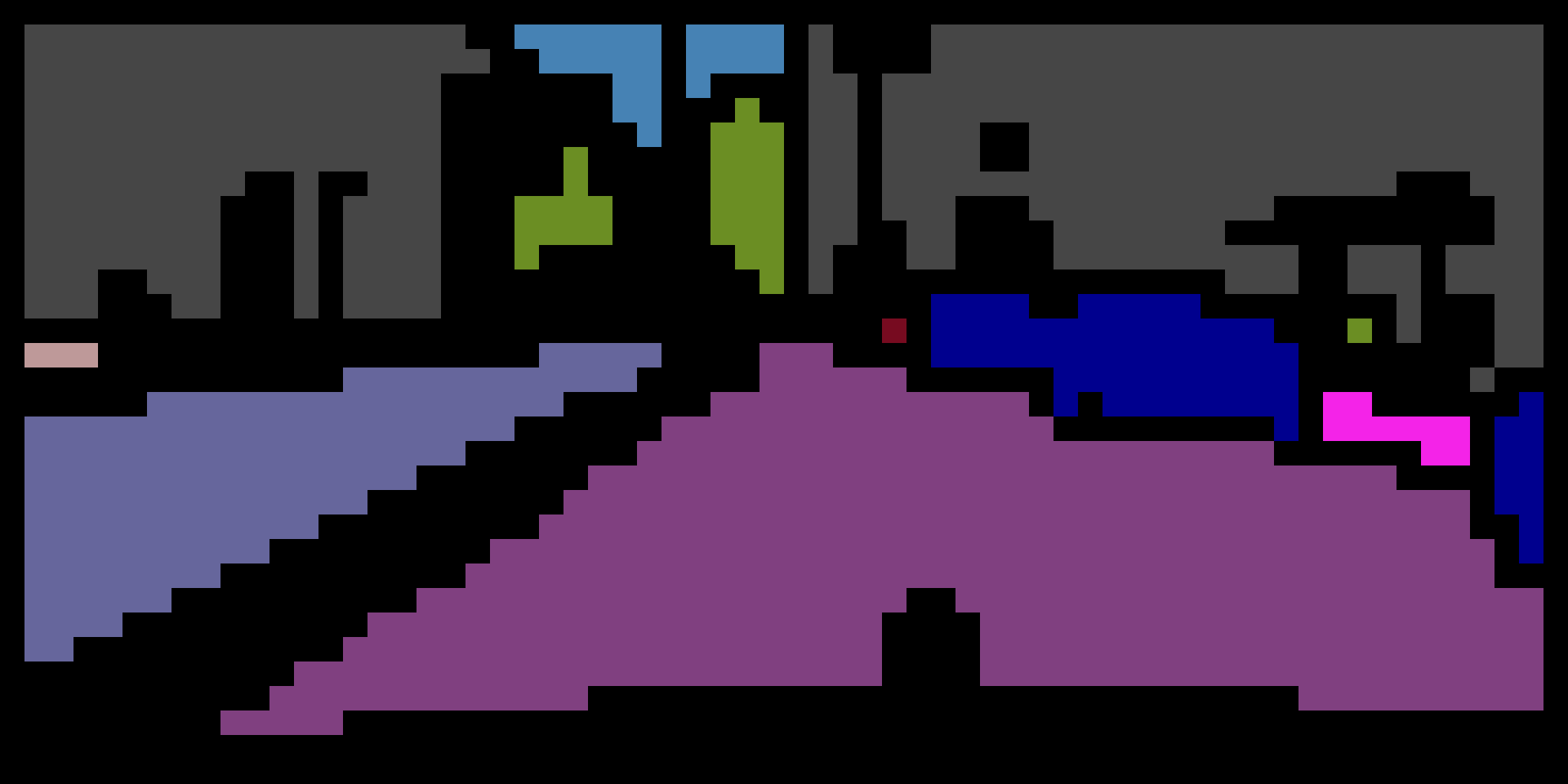} &
        \includegraphics[width=0.19\textwidth]{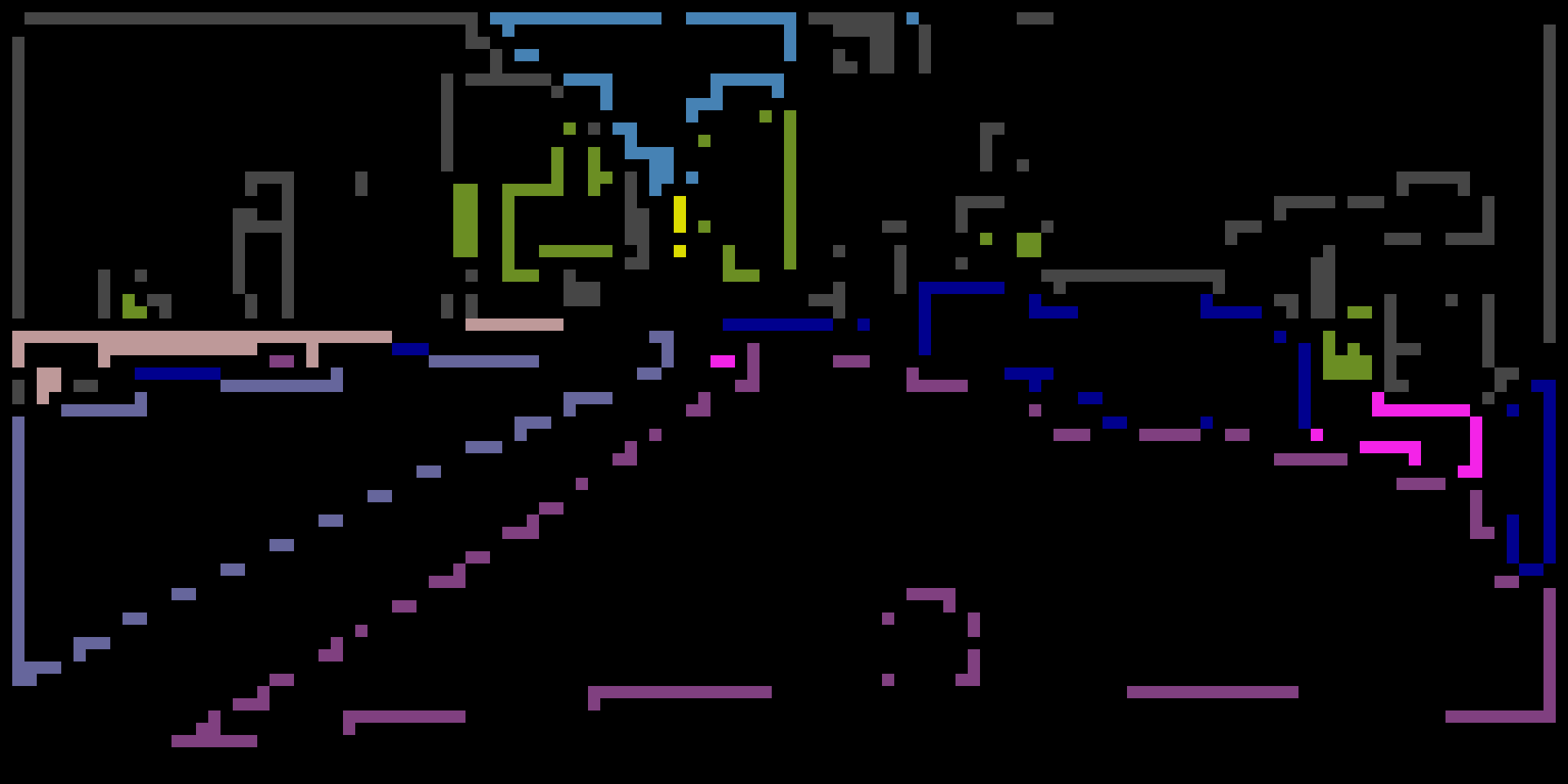} &
        \includegraphics[width=0.19\textwidth]{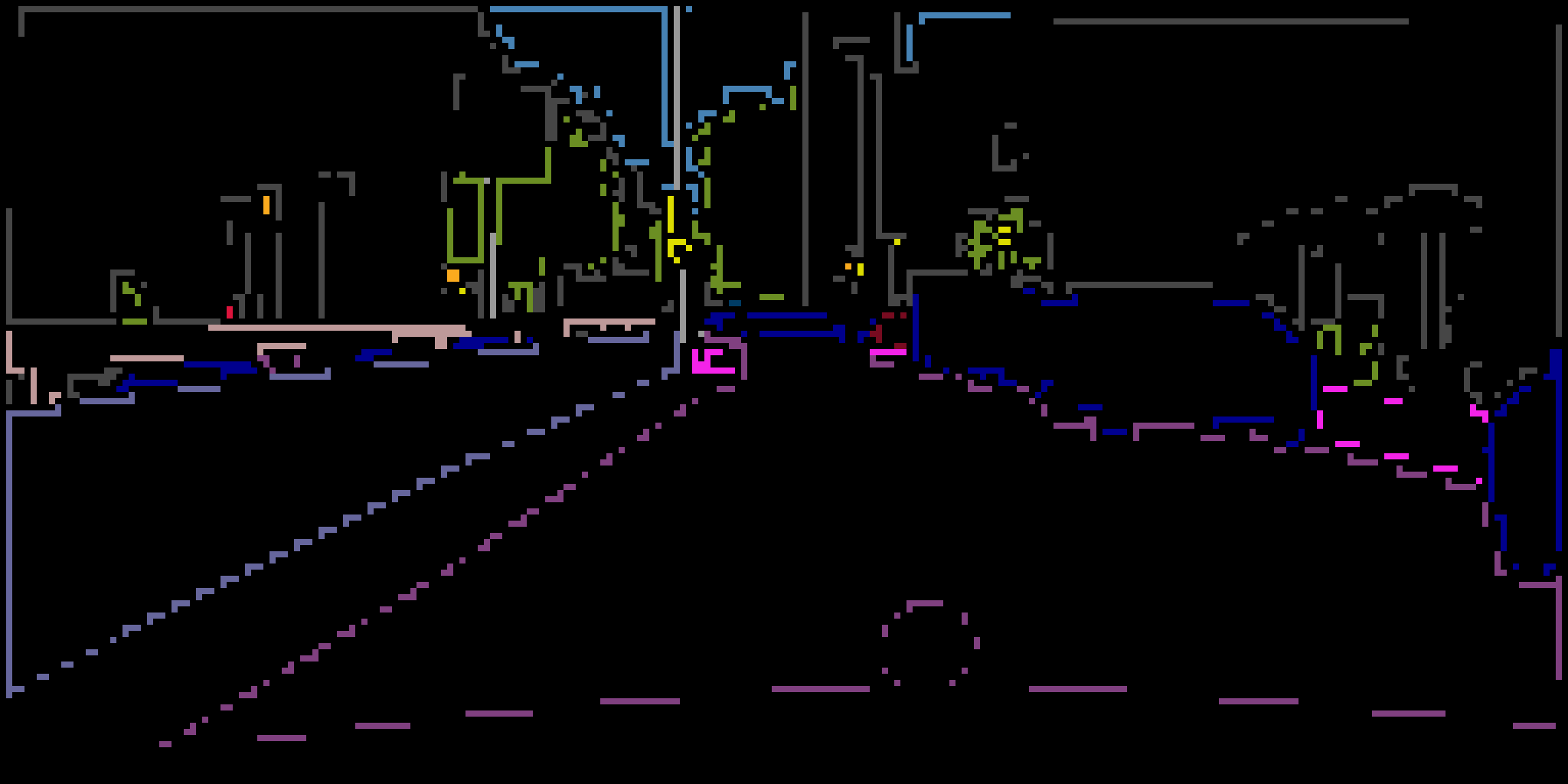} \\
        Segmentation Mask & $Q_2$ & $Q_1$ & $Q_0$\\
        \includegraphics[width=0.19\textwidth]{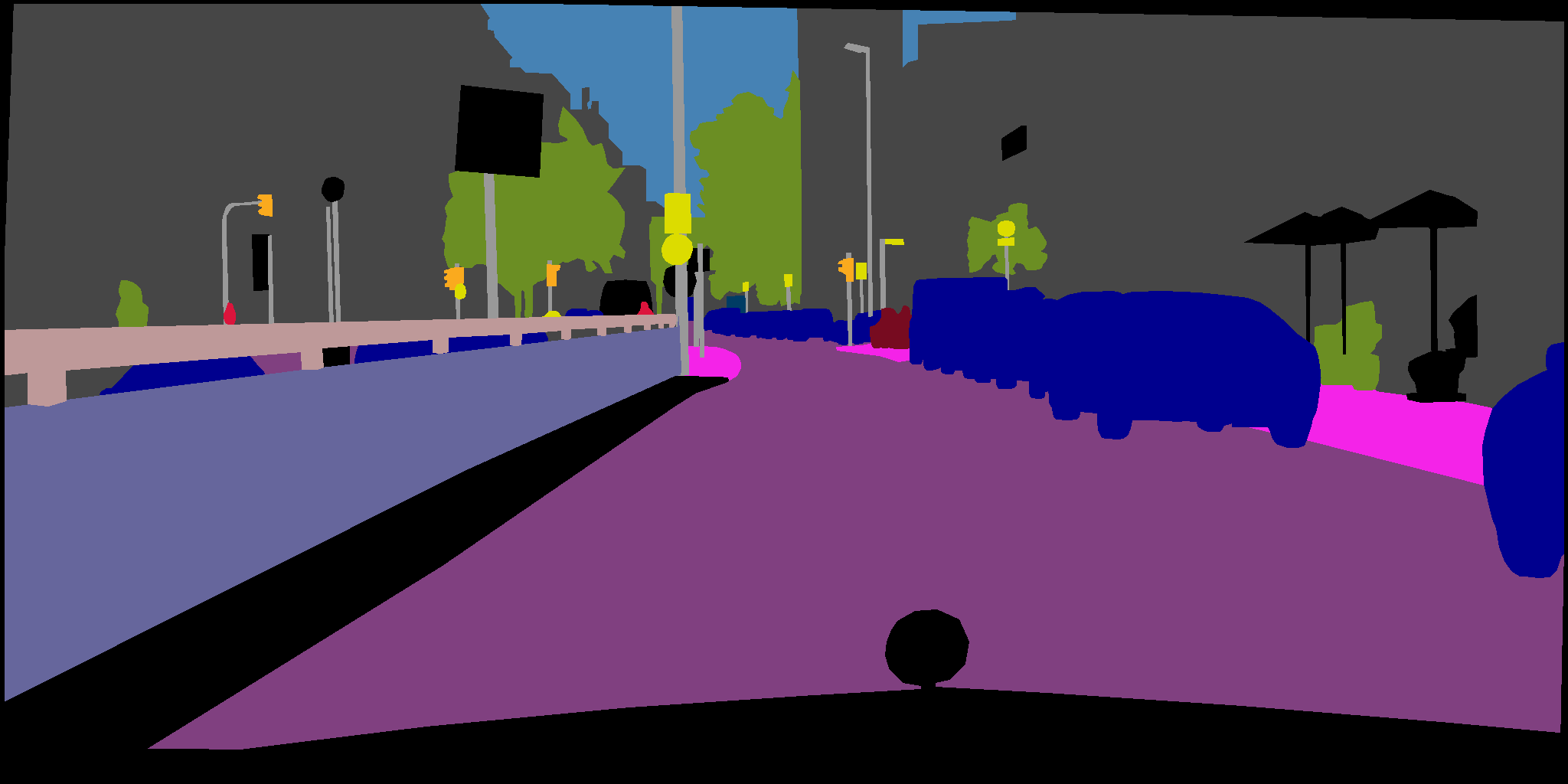} & \includegraphics[width=0.19\textwidth]{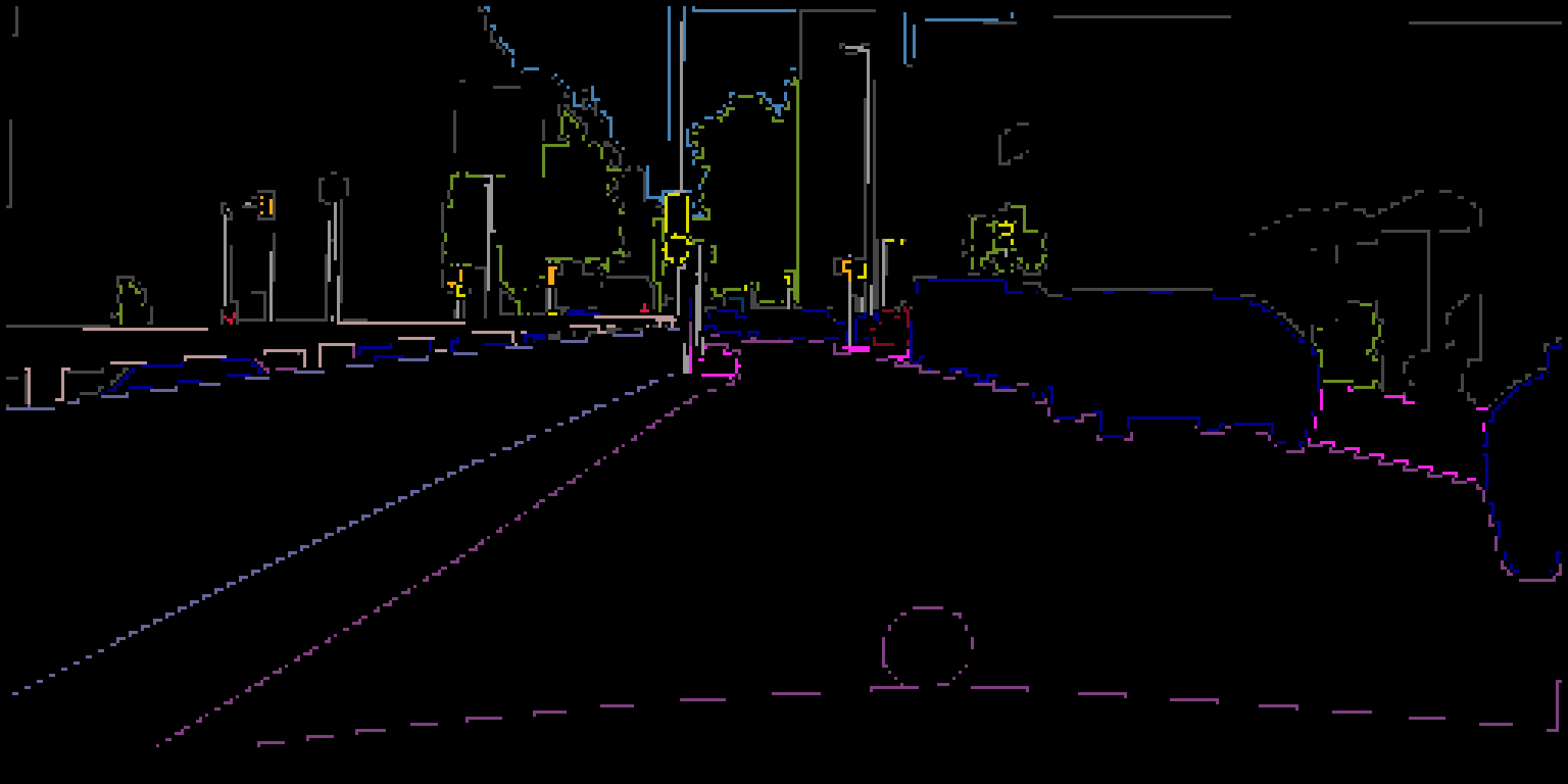} &
        \includegraphics[width=0.19\textwidth]{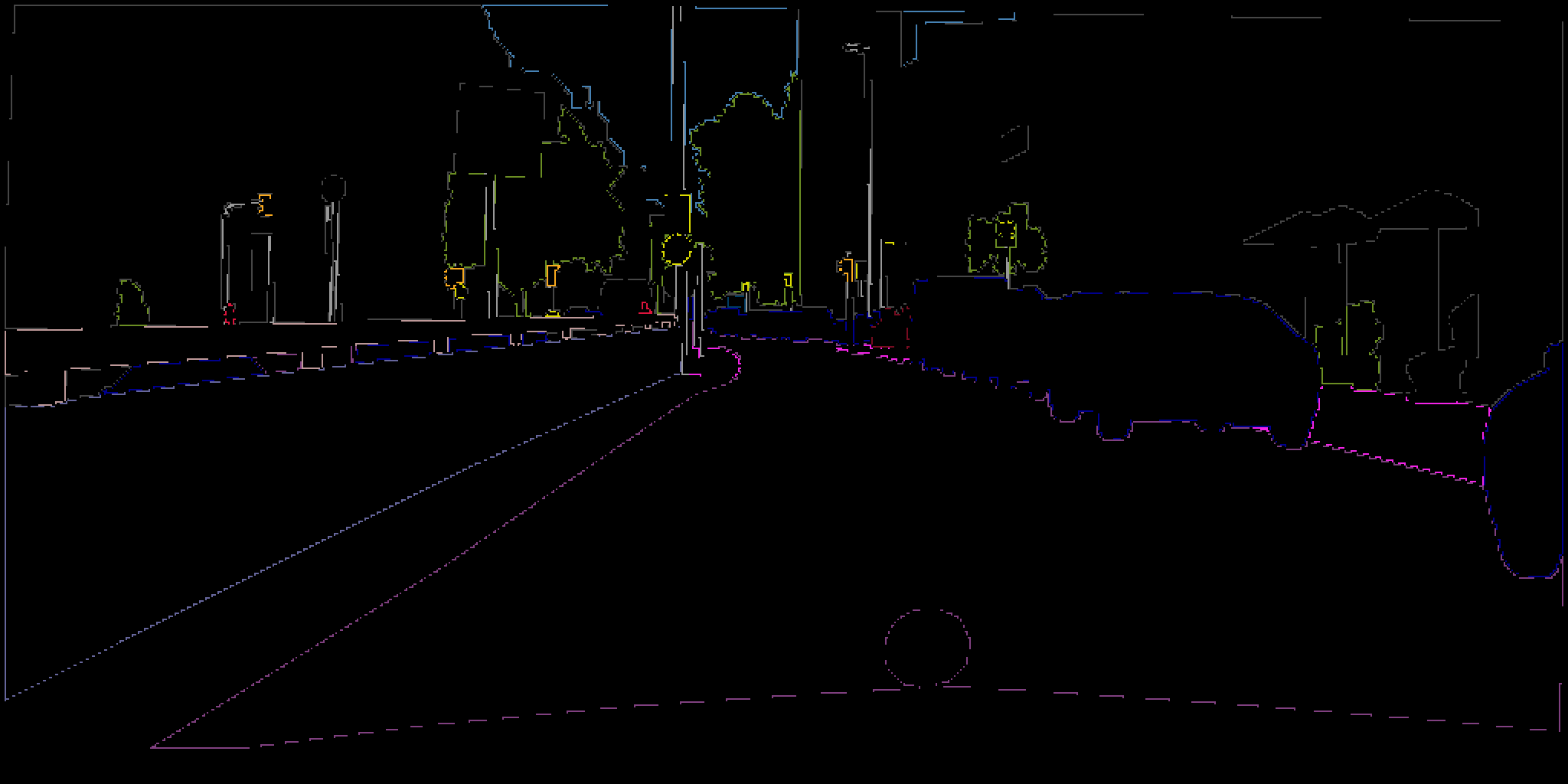} &
        \includegraphics[width=0.19\textwidth]{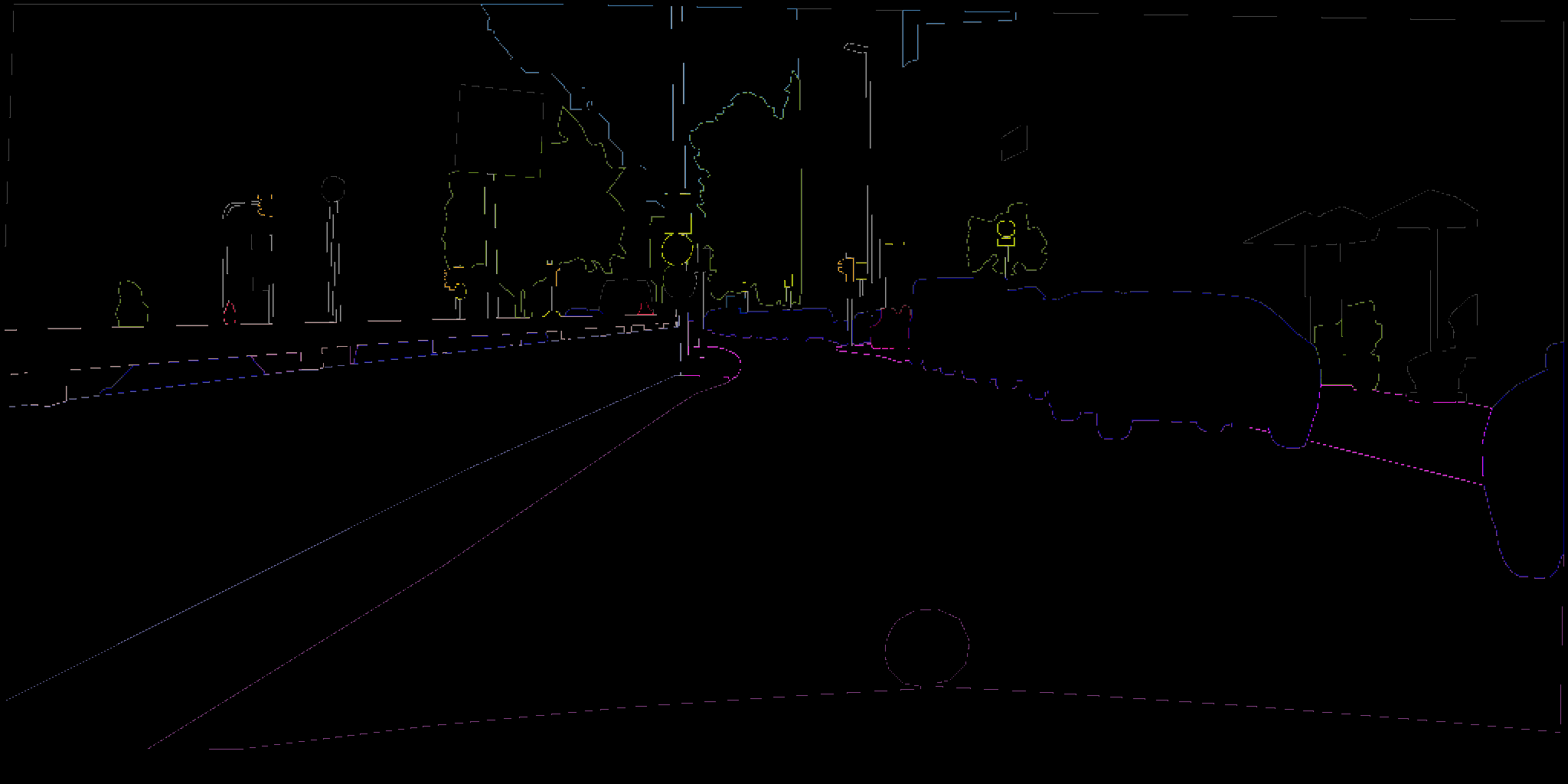} \\
		\hline
		\\
		\hline
		Image & $Q_5$ & $Q_4$ & $Q_3$ \\
		\includegraphics[width=0.19\textwidth]{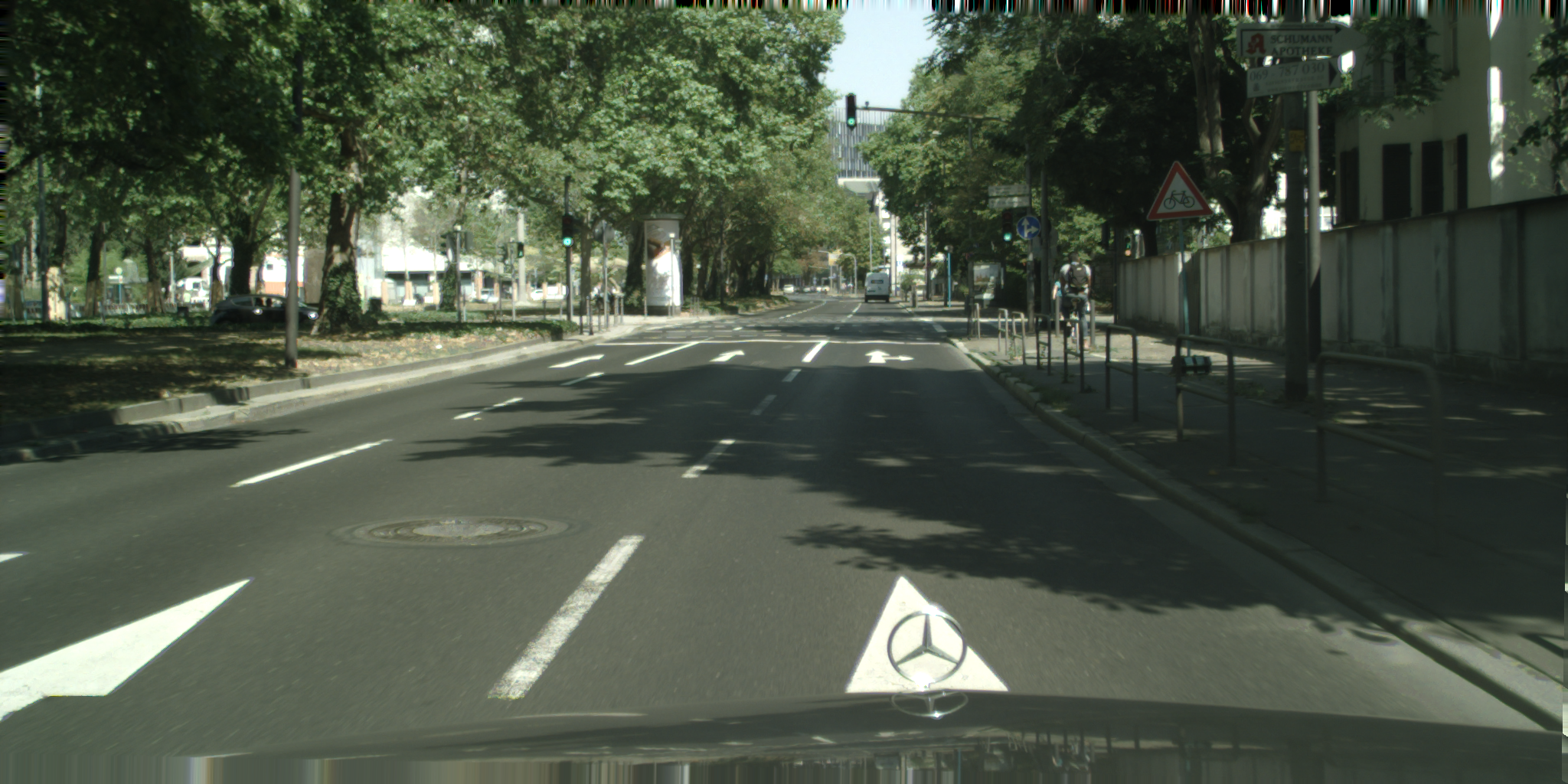} &
		\includegraphics[width=0.19\textwidth]{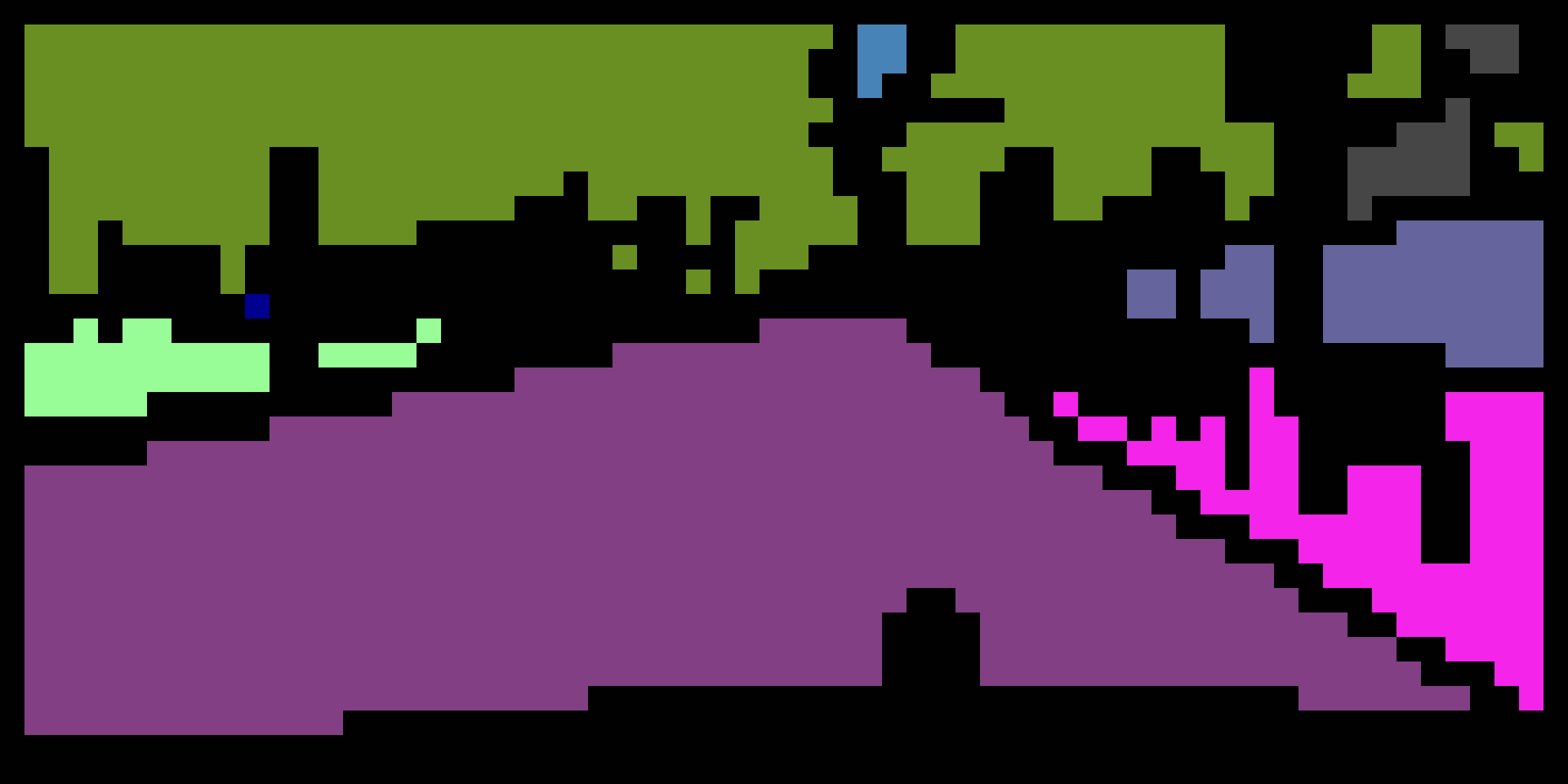} &
        \includegraphics[width=0.19\textwidth]{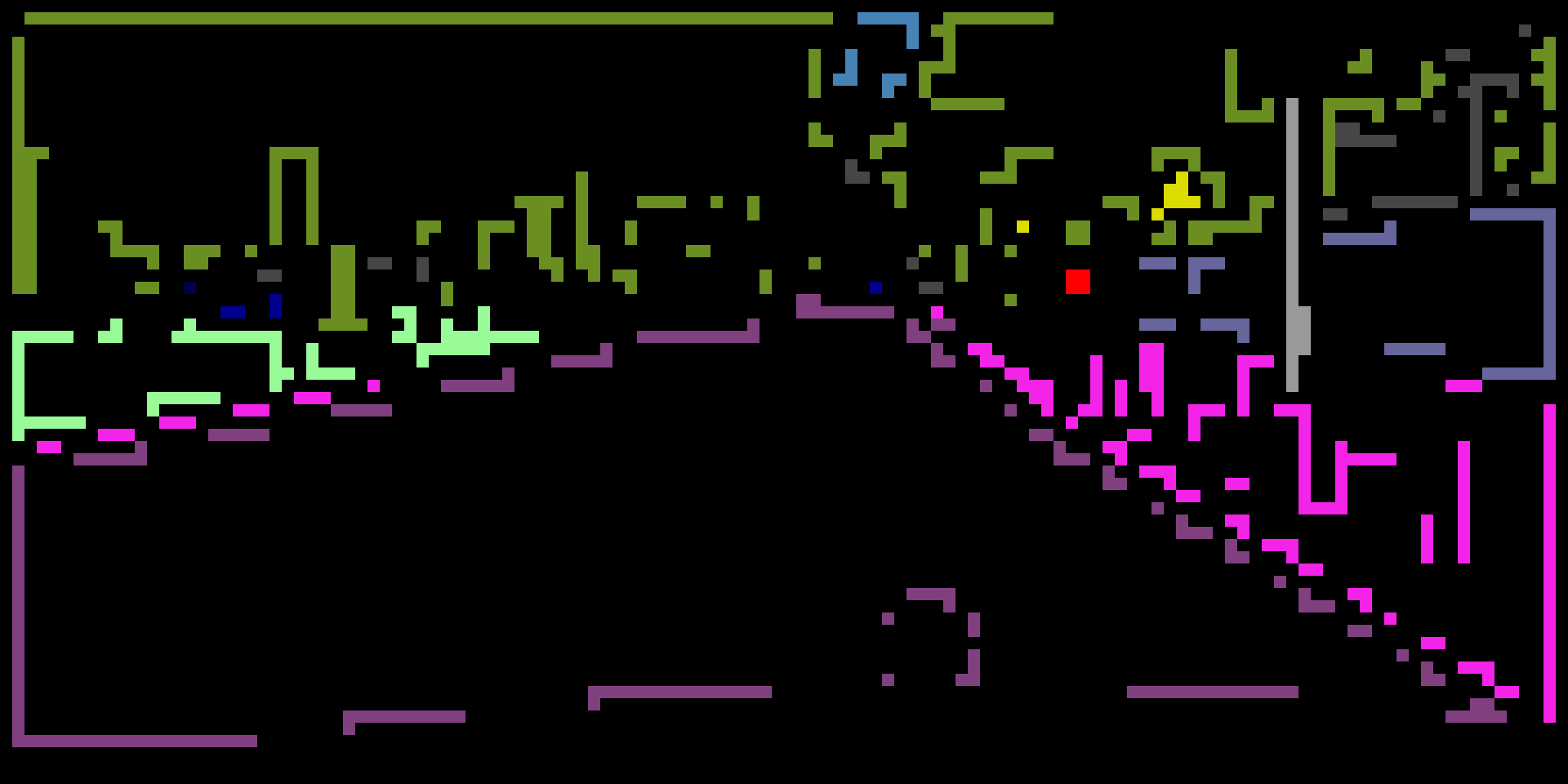} &
        \includegraphics[width=0.19\textwidth]{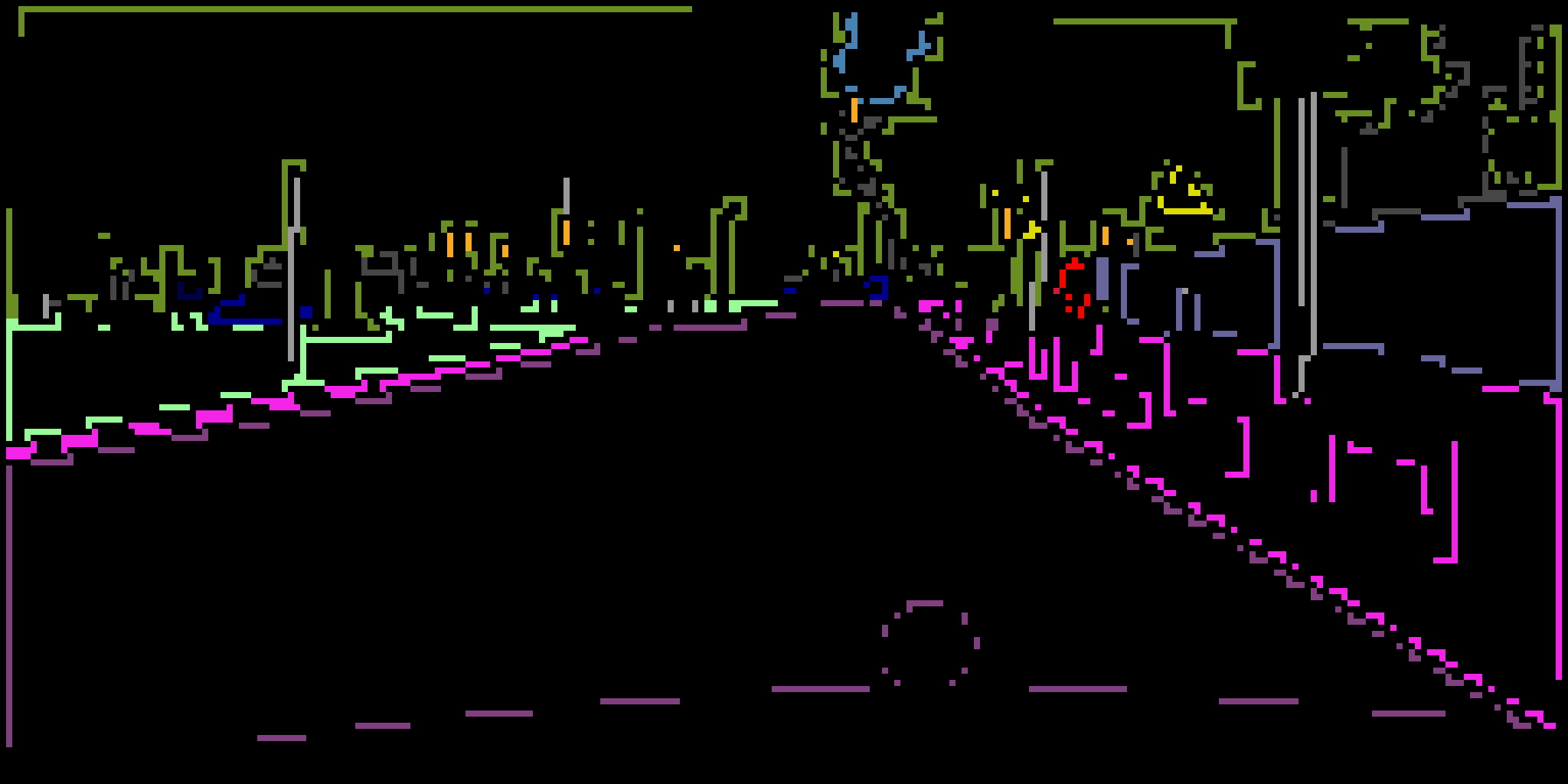} \\
        Segmentation Mask & $Q_2$ & $Q_1$ & $Q_0$\\
        \includegraphics[width=0.19\textwidth]{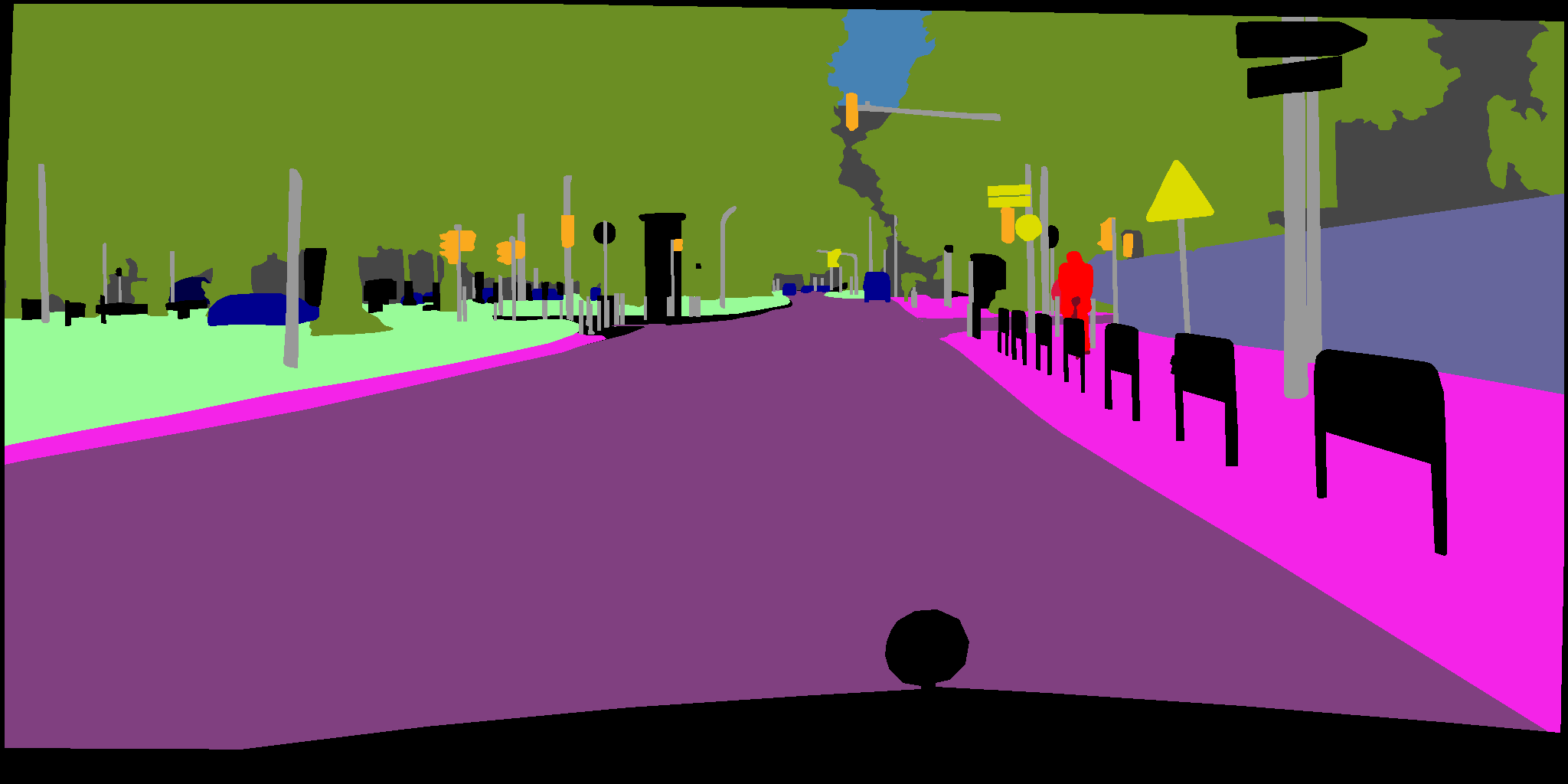} & \includegraphics[width=0.19\textwidth]{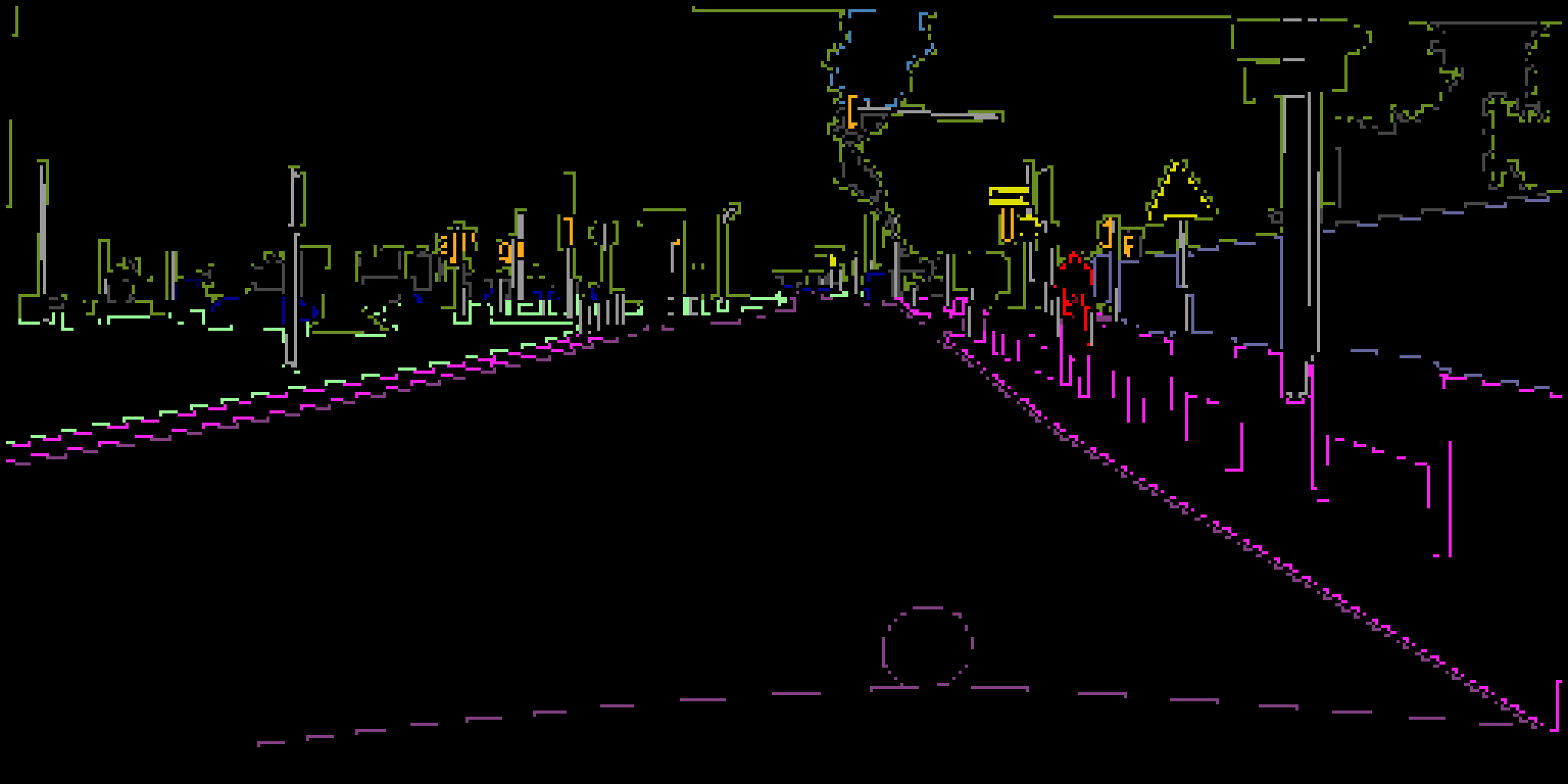} &
        \includegraphics[width=0.19\textwidth]{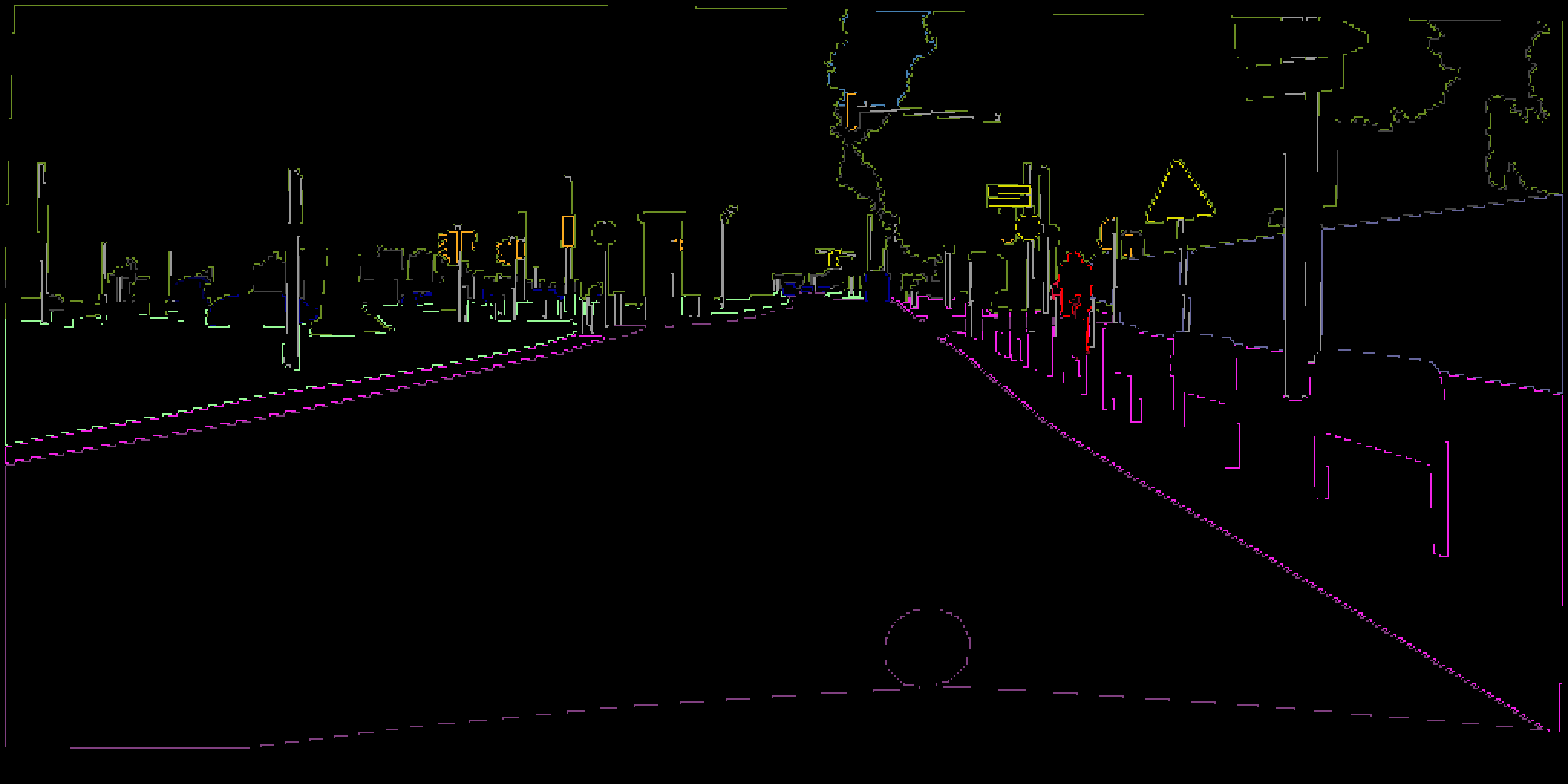} &
        \includegraphics[width=0.19\textwidth]{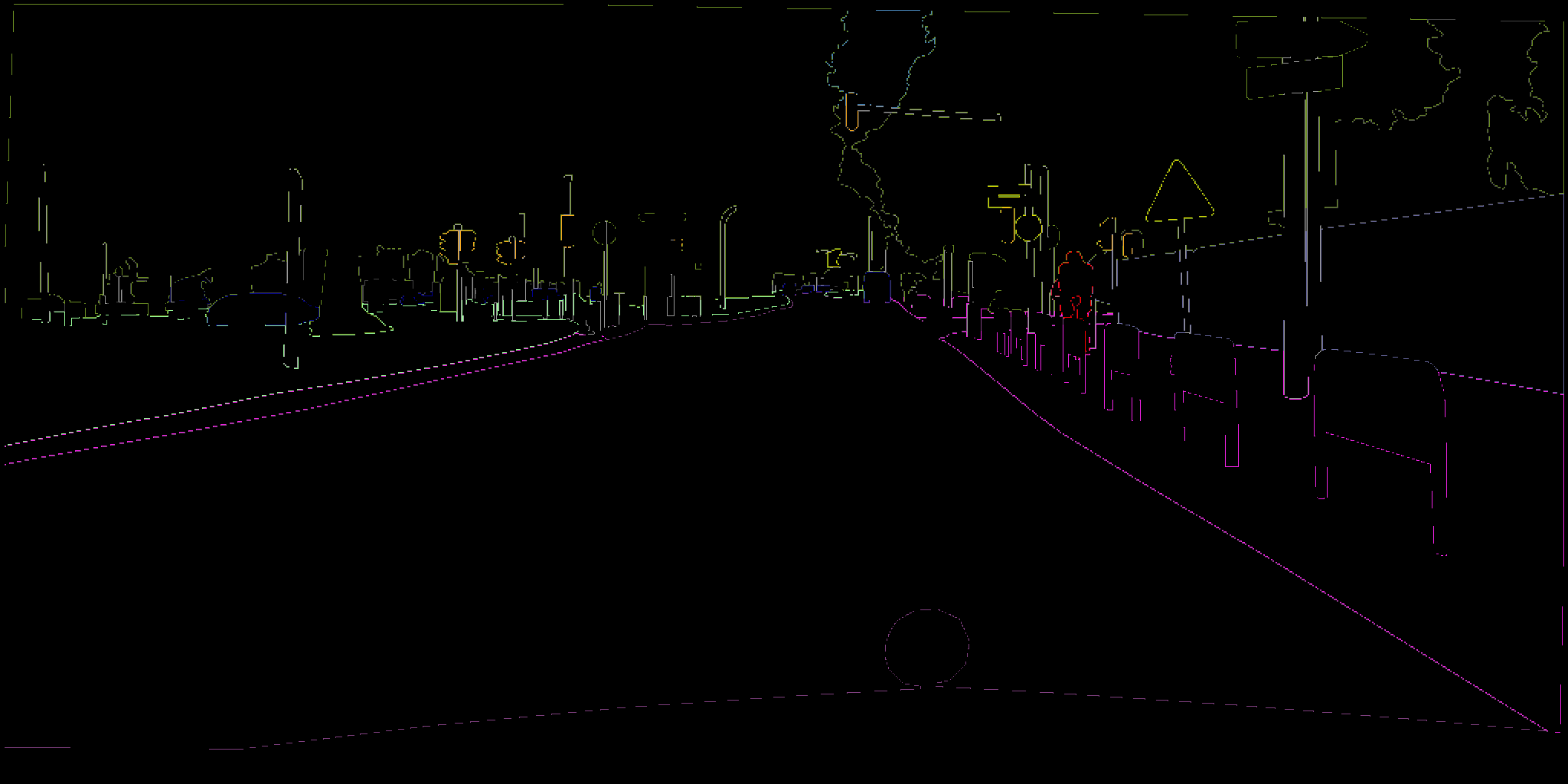} \\
		\hline
	\end{tabular}
	\label{t:city_vis}
\end{sidewaystable}

\begin{table}
\centering
\caption{Visualization of dataset sparsity using examples from the ADE20k validation set. ${Q}_5, ..., {Q}_0$ are the different quadtree levels. As noted in Table \ref{t:datasets}, a large fraction of the pixels belong to larger cells (such as ${Q}_5$), while the leaf ($Q_0$) nodes are extremely sparse.  The composite class is represented in black (best viewed in color).}
\begin{tabular}[\textheight]{c|ccccccc}
		\hline
		Image & $Q_5$ & $Q_4$ & $Q_3$ \\
        \includegraphics[width=0.17\textwidth]{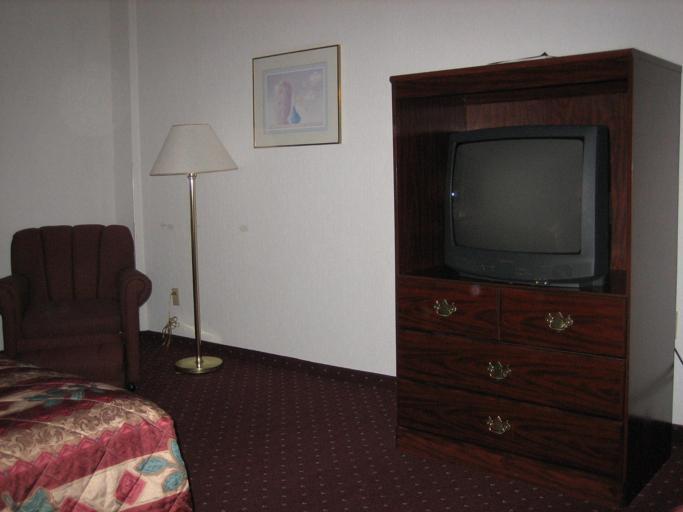} & \includegraphics[width=0.17\textwidth]{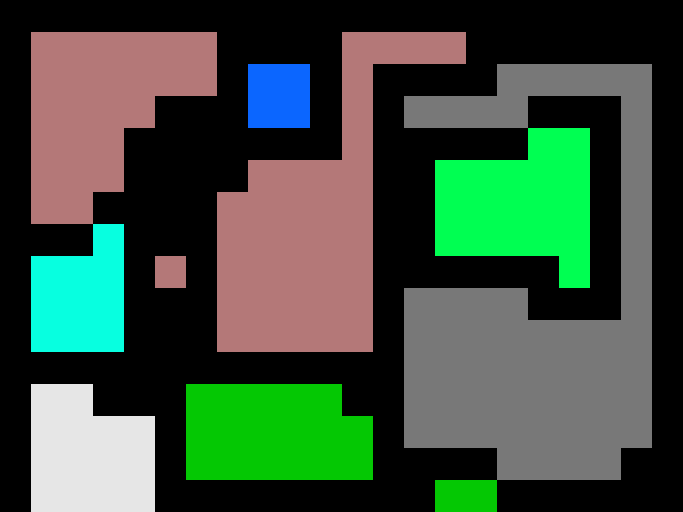} &
        \includegraphics[width=0.17\textwidth]{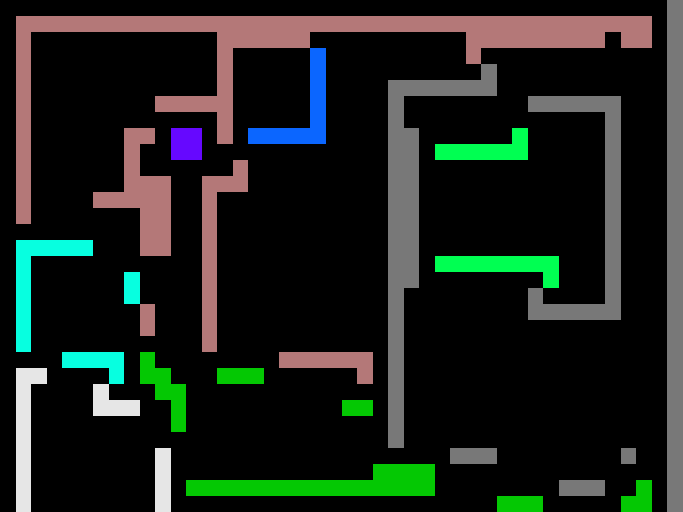} &
        \includegraphics[width=0.17\textwidth]{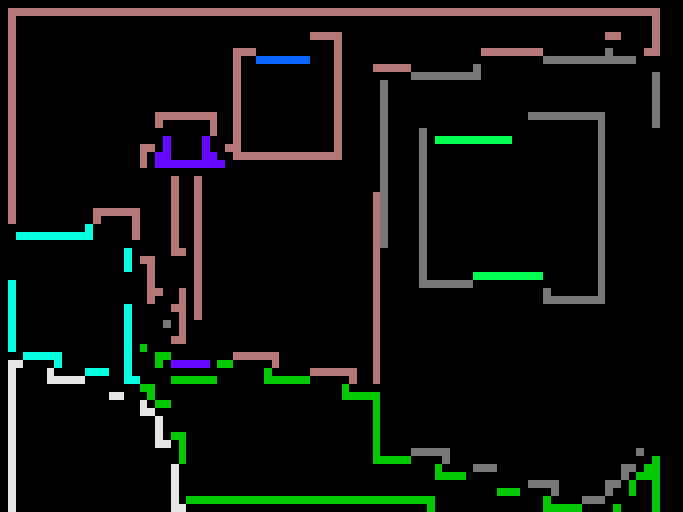} \\
        Segmentation Mask & $Q_2$ & $Q_1$ & $Q_0$\\
        \includegraphics[width=0.17\textwidth]{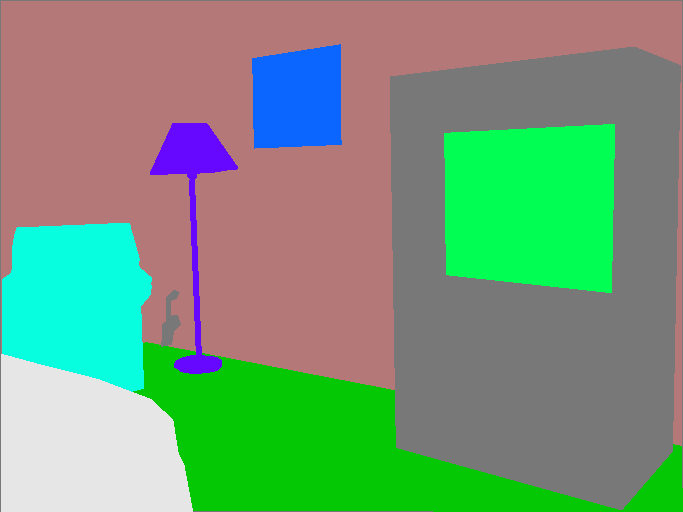} & \includegraphics[width=0.17\textwidth]{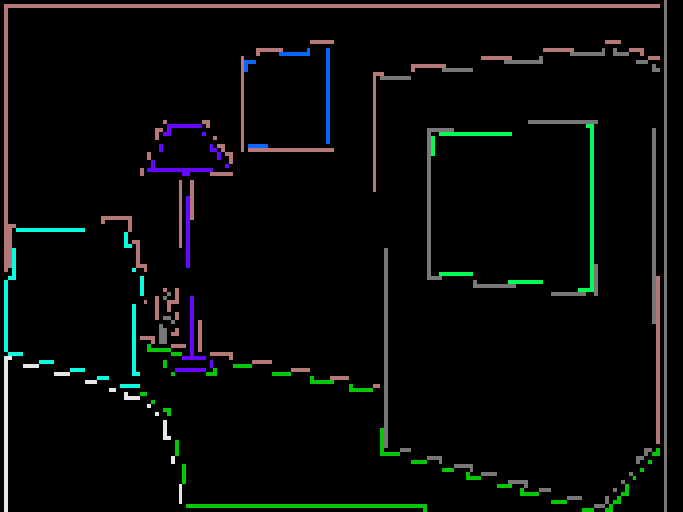} &
        \includegraphics[width=0.17\textwidth]{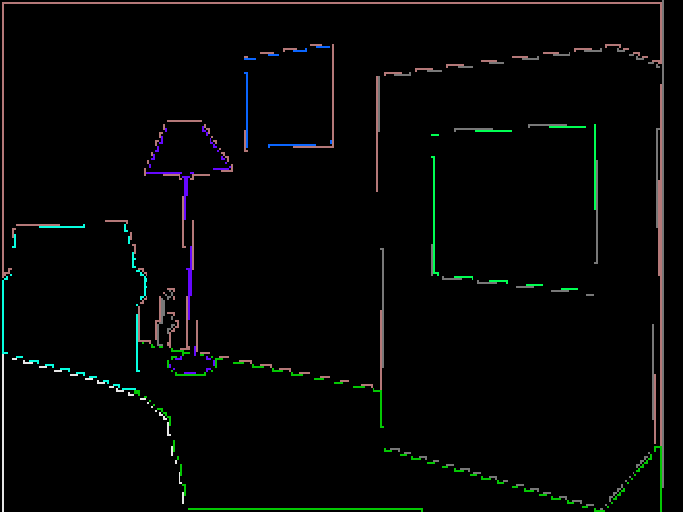} &
        \includegraphics[width=0.17\textwidth]{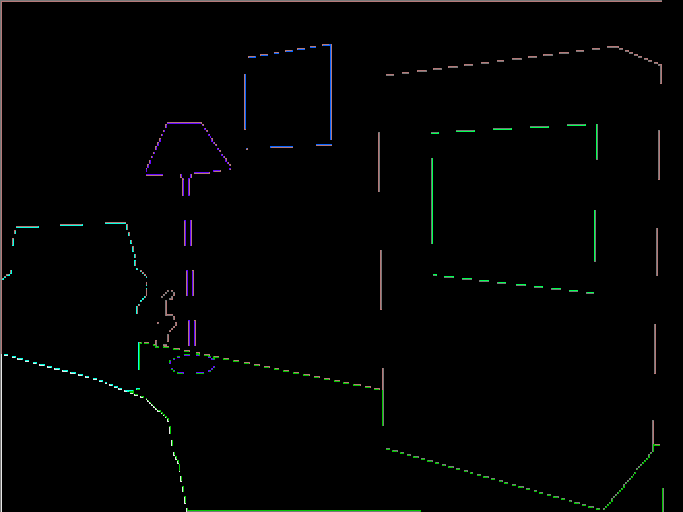} \\
		\hline
		\\
		\hline
		Image & $Q_5$ & $Q_4$ & $Q_3$ \\
		\includegraphics[width=0.17\textwidth]{latex/supp_figs/ADE_val_00001083.jpg} & \includegraphics[width=0.17\textwidth]{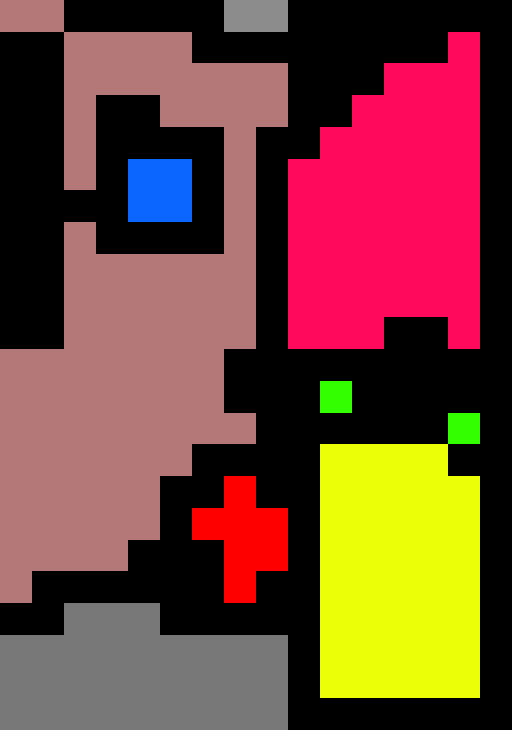} &
        \includegraphics[width=0.17\textwidth]{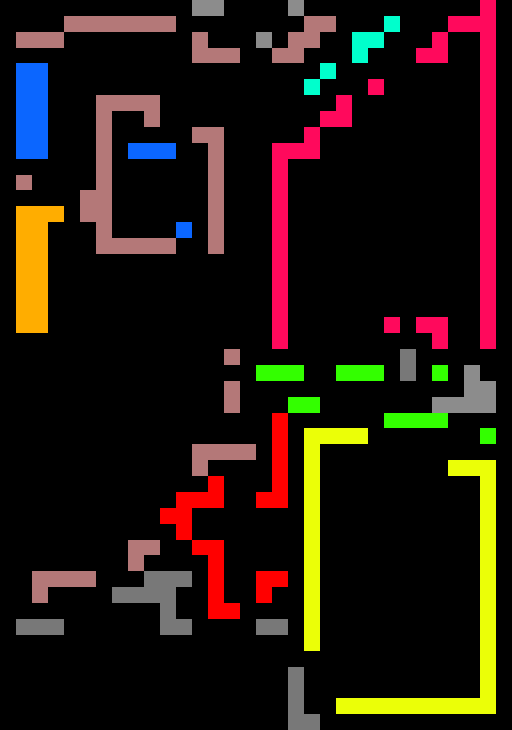} &
        \includegraphics[width=0.17\textwidth]{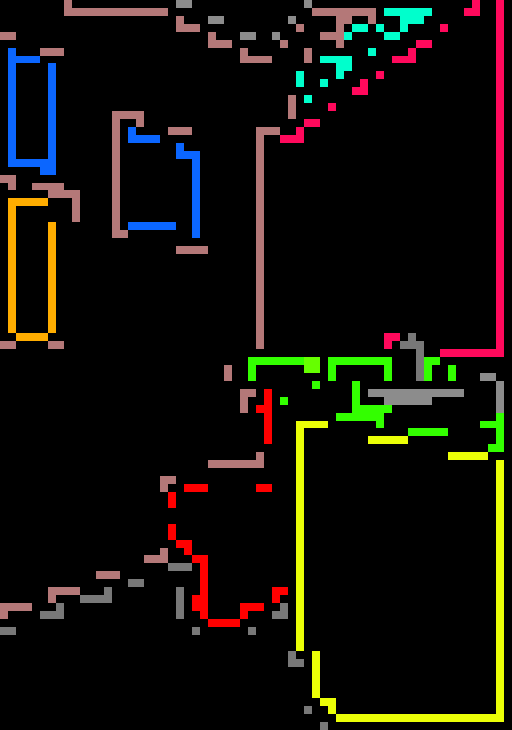} \\
        Segmentation Mask & $Q_2$ & $Q_1$ & $Q_0$\\
        \includegraphics[width=0.17\textwidth]{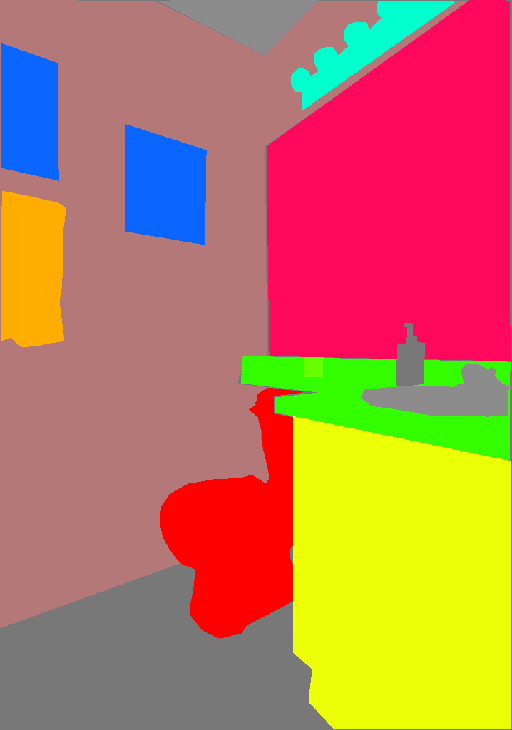} & \includegraphics[width=0.17\textwidth]{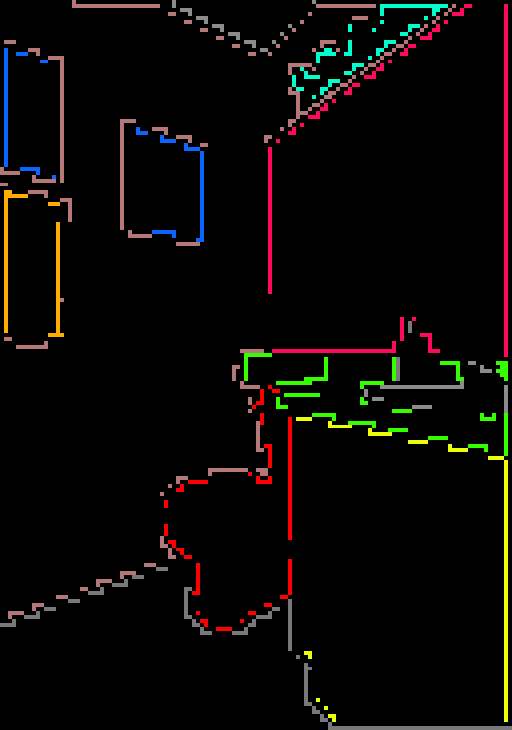} &
        \includegraphics[width=0.17\textwidth]{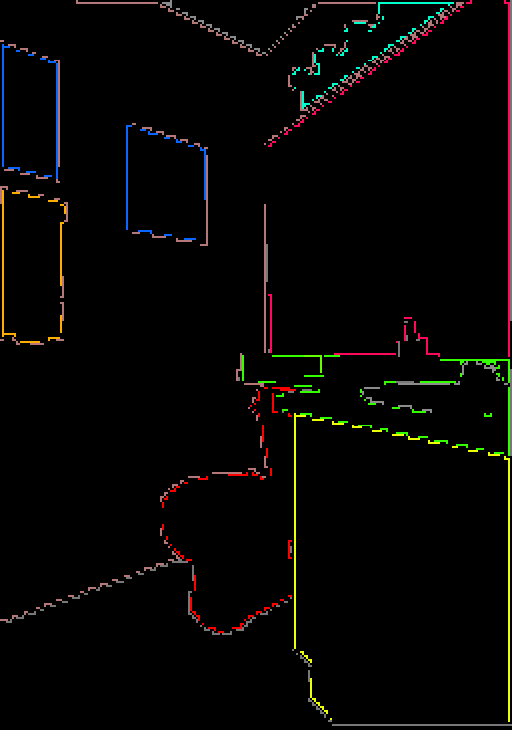} &
        \includegraphics[width=0.17\textwidth]{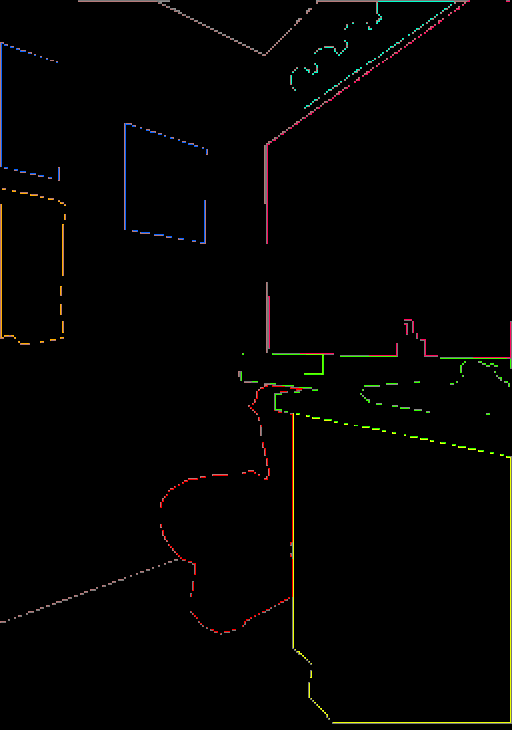} \\
		\hline
		\\
		\hline
		Image & $Q_5$ & $Q_4$ & $Q_3$ \\
		\includegraphics[width=0.17\textwidth]{latex/supp_figs/ADE_val_00001724.jpg} & \includegraphics[width=0.17\textwidth]{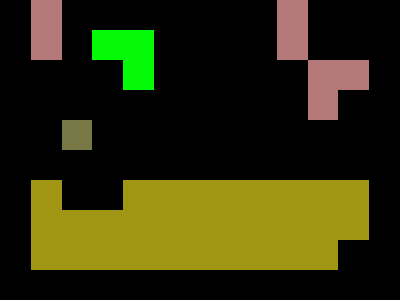} &
        \includegraphics[width=0.17\textwidth]{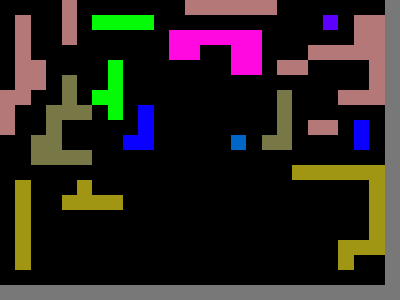} &
        \includegraphics[width=0.17\textwidth]{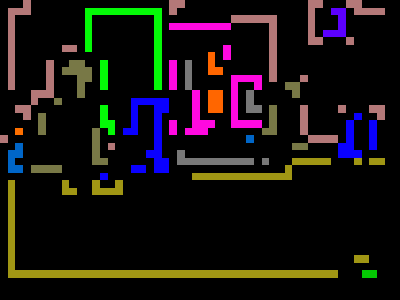} \\
        Segmentation Mask & $Q_2$ & $Q_1$ & $Q_0$\\
        \includegraphics[width=0.17\textwidth]{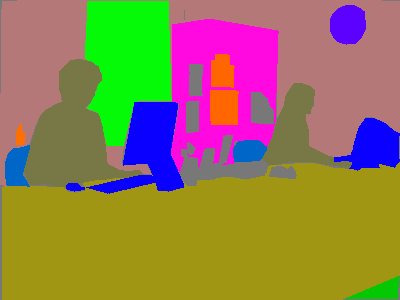} & \includegraphics[width=0.17\textwidth]{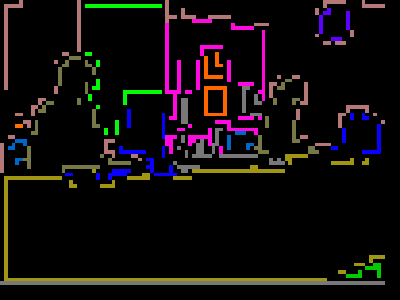} &
        \includegraphics[width=0.17\textwidth]{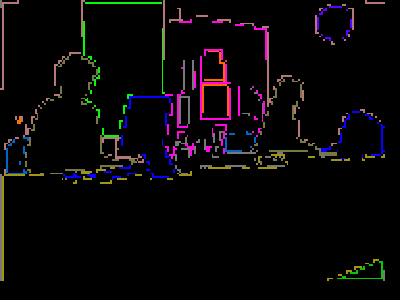} &
        \includegraphics[width=0.17\textwidth]{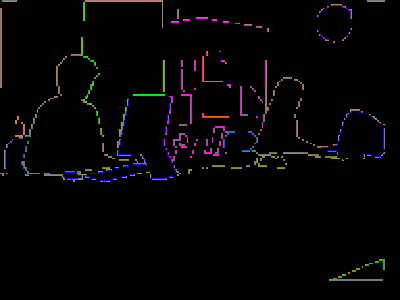} \\
		\hline
	\end{tabular}
	\label{t:ade_vis}
\end{table}

\end{document}